\documentclass{article}

\usepackage{arxiv}
\usepackage{amssymb}
\usepackage{amsmath}
\usepackage{bm}
\usepackage{subfig}
\usepackage{wrapfig}
\usepackage{lipsum}     
\usepackage{float} 
\usepackage{graphicx}           
\usepackage{url} 
\usepackage{soul}
\usepackage{lineno}
\usepackage{diagbox}
\usepackage{color}
\usepackage{enumitem}

\usepackage[ruled,vlined,linesnumbered,noresetcount]{algorithm2e}

\title{Gaussian processes meet NeuralODEs: A Bayesian framework for learning the dynamics of partially observed systems from scarce and noisy data}

\author{
  Mohamed Aziz Bhouri \\
  Department of Mechanical Engineering \\
  and Applied Mechanics\\
  University of Pennsylvania\\
  Philadelphia, PA 19104 \\
  \texttt{bhouri@seas.upenn.edu} \\
   \And
  Paris Perdikaris \\
  Department of Mechanical Engineering \\
  and Applied Mechanics\\
  University of Pennsylvania\\
  Philadelphia, PA 19104 \\
  \texttt{pgp@seas.upenn.edu} \\
}

\begin{document}
\maketitle

\begin{abstract}

This paper presents a machine learning framework (GP-NODE) for Bayesian systems identification from partial, noisy and irregular observations of nonlinear dynamical systems. The proposed method takes advantage of recent developments in differentiable programming to propagate gradient information through ordinary differential equation solvers and perform Bayesian inference with respect to unknown model parameters using Hamiltonian Monte Carlo sampling and Gaussian Process priors over the observed system states. This allows us to exploit temporal correlations in the observed data, and efficiently infer posterior distributions over plausible models with quantified uncertainty. Moreover, the use of sparsity-promoting priors such as the Finnish Horseshoe for free model parameters enables the discovery of interpretable and parsimonious representations for the underlying latent dynamics. A series of numerical studies is presented to demonstrate the effectiveness of the proposed GP-NODE method including predator-prey systems, systems biology, and a 50-dimensional human motion dynamical system. Taken together, our findings put forth a novel, flexible and robust workflow for data-driven model discovery under uncertainty. All code and data accompanying this manuscript are available online at \url{https://github.com/PredictiveIntelligenceLab/GP-NODEs}.

\end{abstract}

\keywords{Scientific machine learning; Dynamical systems; Uncertainty quantification; Model discovery}

%% main text
\section{Introduction}\label{sec:intro}

The task of learning dynamical systems from data is receiving increased attention across diverse scientific disciplines including bio-mechanics \cite{Wang2008}, bio-medical imaging \cite{kak2002principles}, biology \cite{ruoff2003temperature}, physical chemistry \cite{feinberg1974dynamics}, climate modeling \cite{palmer1999nonlinear}, and fluid dynamics \cite{raissi2020hidden}. In order to build a better physical understanding of a dynamical system's behavior, it is crucial to identify interpretable underlying features that govern its evolution \cite{haller2002lagrangian}. The latter enables the possibility of providing reliable future forecasts, and, as a consequence, the ability to improve intervention strategies \cite{tantet2018crisis} via appropriate design or control decisions \cite{bemporad1999control, lu2019nonparametric}.

The evolution of physical systems can be often characterized by differential equations, and several data-driven techniques have been recently developed to synthesize observational data and prior domain knowledge in order to learn dynamics from time-series data \cite{rackauckas2020universal,gholami2019anode,chen2018neural,brunton2016discovering,rudy2017data, brennan2018data,raissi2018multistep, qin2019data, bertalan2019learning, rico1993continuous, gonzalez1998identification}, infer the solution of differential equations \cite{raissi2017inferring, raissi2019physics, zhu2019physics, yang2019adversarial, yang2020physics}, infer parameters, latent variables and unknown constitutive laws \cite{wang2019non, raissi2020hidden, raissi2018hidden, tartakovsky2018learning}, as well as tackle forward and inverse problems for complex dynamical systems including cardiovascular flow dynamics \cite{kissas2020machine}, meta-materials \cite{chen2019physics}, and epidemiology models \cite{Bhouri2020}.

Most existing data-driven systems identification techniques heavily depend on the quality of the available observations and do not provide uncertainty quantification \cite{rackauckas2020universal, brunton2016discovering, qin2019data, chen2018neural,gholami2019anode}. For instance, the sparse regression methods put forth in \cite{brunton2016discovering, rudy2017data} cannot be directly applied to irregularly sampled time-series or to trajectories with missing values, and can become unstable in the presence of highly noisy data. More recent approaches leverage differentiable programming to construct continuous-time models such as NeuralODEs \cite{rackauckas2020universal,chen2018neural,gholami2019anode}, and are able to accommodate observations with irregular sampling frequency, but they are not designed to provide a predictive uncertainty of the discovered dynamics. The latter has been recently enabled by the Bayesian formulations of Yang {\it et al.} \cite{yang2020bayesian} and Dandekar {\it et al.} \cite{dandekar2020bayesian}. However, as discussed in \cite{yang2020bayesian}, these methods typically require tuning to obtain a good initialization for a Markov Chain Monte Carlo (MCMC) sampler, while the use of Laplace priors cannot consistently promote sparsity in the underlying dynamics, unless ad-hoc thresholding procedures are employed. Moreover, most aforementioned approaches typically assume that the initial conditions are known and all state variables can be observed. These limitations prevent applying such techniques to non-ideal settings where data is noisy, sparse and irregularly sampled, the initial conditions may be unknown, and/or some of the system variables are latent (i.e. cannot be observed).

In order to build a systems identification framework that alleviates difficulties related to low frequency and regularity in the observed data, one can employ Gaussian process (GP) priors to model temporal correlations in observed state variables \cite{Pillonetto2014, Calderhead2009,Heinonen2018,Wenk2019,Hamzi2021}. One appealing feature of GPs is their ability to automatically enforce the ``Occam's razor" principle \cite{MacKay1992}, leading to increased robustness against over-fitting in the small data regime, assuming that the underlying dynamics is sufficiently smooth. However most aforementioned techniques either require a perfectly known form of the dynamical system, which prevents the possibility of discovering  unknown model forms, or model the whole dynamical system as a Gaussian process, which does not typically yield an interpretable representation of the latent dynamics.

% Specifically, the log-determinant term in the GP likelihood is essentially a regularizer that favors finding the simplest possible model that best fits the observed data, which alleviate the burden of model selection. 
% Recently, several works have been proposed to embed parametric differential equations with Gaussian process frameworks \cite{Calderhead2009,Heinonen2018,Wenk2019,Hamzi2021}, showing the efficiency of the latter given that the underlying dynamics is sufficiently smooth. However these techniques either require a perfectly known form of the dynamical system,  which prevents the possibility of discovering the unknown model form, or model the whole dynamical system as a Gaussian process, which does not typically yield an interpretable representation of the latent dynamics.

In this work we employ Gaussian process priors and automatic differentiation through ODE solvers to formulate a Bayesian framework (GP-NODE) for dynamical systems identification under scarce, noisy and partially observed data. Moreover, we employ a consistent Bayesian treatment for promoting sparsity in latent dynamics by employing the Horseshoe and Finnish Horseshoe priors \cite{Carvalho2009,piironen2017a,Piironen2017b}. This enables us to seamlessly tackle cases involving a known model parametrization, or cases for which the underlying model form is unknown but can be represented by a large dictionary. 
The novel capabilities of the resulting computational framework can be summarized in the following points:
\begin{itemize}[leftmargin=*]
\setlength\itemsep{0em}
    \item Leveraging GP priors, the proposed GP-NODE workflow can naturally model temporal correlations in the data and remain robust in the presence of scarce and noisy observations, latent variables, unknown initial conditions, irregular time sampling for each observable variable, and observations at different time instances for the observable variables. 
    \item  A key element of our framework is the design of novel GP mean functions with excellent extrapolation performance using classical ODE integrators (e.g. Runge-Kutta methods).
    \item  Leveraging automatic differentiation, we develop an end-to-end differentiable pipeline that enables the use of state-of-the-art Hamiltonian Monte Carlo samplers for accelerated Bayesian inference (e.g. NUTS \cite{hoffman2014no}).
    \item Thanks to sparsity-promoting Finnish Horseshoe priors, the proposed approach can recover interpretable and parsimonious representations of the latent dynamics.
    % \item This probabilistic formulation is robust against erroneous data and produces reliable future forecasts with uncertainty quantification.
    % \item We demonstrate enhanced capabilities and robustness against state-of-the-art methods for systems identification \cite{brunton2016discovering, rudy2017data, Heinonen2018} across a range of benchmark problems.
\end{itemize}
Taken all together, our findings put forth a novel, flexible and robust workflow for data-driven model discovery under uncertainty that can potentially lead to improved algorithms for forecasting, control and model-based reinforcement learning of complex systems.

The rest of this paper is organized as follows. Section \ref{sec:methods} presents the GP-NODE method and the corresponding technical ingredients needed to perform Bayesian inference using Gaussian processes with NeuralODE mean functions. Two different problem settings are considered: parameter inference with an unknown model form and parameter inference with domain knowledge incorporation. The data normalization step is also presented within section \ref{sec:methods} in order to obtain a universal model initialization and unified prior distributions across the various numerical examples presented in section \ref{sec:Results}. These numerical studies include dynamics discovery for predator-prey system with an unknown model form, including extensive comparisons against the SINDy algorithm of Brunton {\em et. al.} \cite{brunton2016discovering}. We also present examples on the calibration of a realistic Yeast Glycolysis model, as well as learning a 50-dimensional dynamical system from motion capture data, for which we also present comparisons against the Nonparametric ODE Model (npODE) of Heinonen {\it et al.} \cite{Heinonen2018}. Finally, in section \ref{sec:discussion} we summarize our key findings, discuss the limitations of the proposed GP-NODE approach, and carve out directions for future investigation.

\section{Methods}\label{sec:methods}
This section provides a comprehensive overview of the key ingredients that define the proposed framework (GP-NODE), namely differential programming with NeuralODEs \cite{chen2018neural} and Bayesian inference with Hamiltonian Carlo sampling and Gaussian process priors \cite{neal2011mcmc, rasmussen2006gaussian}. Our presentation is focused on describing how these techniques can be interfaced to obtain an efficient and robust workflow for model discovery from imperfect time-series data.

\subsection{GP-NODE: Gaussian processes with NeuralODE mean functions}\label{sec:diff_prog}

We consider a general dynamical system of the form:

\begin{equation}\label{equ:dynamics}
    \frac{d \bm{x}}{dt} = f(\bm{x}, t; \bm{\theta_f}),
\end{equation}
where $\bm{x}\in\mathbb{R}^{D}$ denotes the $D$-dimensional state space, and $\bm{\theta_f}$ is a vector unknown parameters that defines the latent dynamics $f:\mathbb{R}^{D}\rightarrow\mathbb{R}^{D}$, and the (potentially) unknown initial conditions. 

Let $V$ be the set of indices of observable variables for which we assume we have data, such that for $d\in V$, we assume that we have observations $\hat{\bm{x}}_d(t^{(d)}_i)$ for $i=1,\ldots,n^{(d)}$; while for $d\notin V$, only the initial condition $\hat{\bm{x}}_d(t=0)$ are observed. Let $D_v\leq D$ denote the size of $V$ and $\bm{t}^{d}\in\mathbb{R}^{n^{(d)}}$ the vector containing $t_i^{(d)}$, for $i=1,\ldots,n^{(d)}$ and $d\in V$. Let $N_v=\sum_{d\in V}n^{(d)}$ and $\bm{t}$ denote the concatenation of $\bm{t}^d$, for $d\in V$.

Let $\bm{z}(t;\bm{\theta_f})$ denote the solution of equation (\ref{equ:dynamics}) at time $t$ for the trainable parameters $\bm{\theta_f}$, obtained using an differentiable ODE solver \cite{chen2018neural},
\begin{equation}\label{eq:NeuralODE}
\bm{z}(t;\bm{\theta_f}) = {\rm ODEsolve}(\bm{z}_0,t,\bm{\theta_f}) \ ,
\end{equation}
\noindent where $\bm{z}_0$ refers to the initial conditions which are either known, or part of the trainable parameters $\bm{\theta_f}$.

Efficiency of using Gaussian processes (GPs) as a prior for dynamics learning from sparse data has been shown in previous studies as long as the underlying dynamics is sufficiently smooth, which can be guaranteed if the observed data are correlated in time \cite{Pillonetto2014,Calderhead2009,Heinonen2018}. Dynamical systems governed by ordinary differential equations fall within such category motivating the choice of using GPs as a prior for the system variables. For $d\in V$, the $d$-th dimension of the state space $\bm{x}$ is modeled as a sum of $Q_d$ uncorrelated GPs, with a mean equal to the $d$-th dimension of $\bm{z}(\bm{\theta_f})$, and covariance functions  $k_{d,q}(\cdot,\cdot,\bm{\theta}_{d,q})$, $q=1,\ldots,Q_d$ such that,
\begin{equation}\label{eq:GP_NeuralODE}
\bm{x_V}(\bm{t})\sim\mathcal{N}(\bm{z_V}(\bm{t};\bm{\theta_f}),\bm{K}(\bm{t},\bm{t},\bm{\theta_g})+\bm{K_\epsilon}(\bm{\theta_g})) \ ,
\end{equation}
\noindent where $\bm{x_V}(\bm{t})\in\mathbb{R}^{N_v}$ is the concatenation of the predictions for the variables $\bm{x}_d$ at the time instances $\bm{t}^{(d)}$ for $d\in V$, $\bm{z_V}(\bm{t})\in\mathbb{R}^{N_v}$ the concatenation of $\bm{z}_d$ at the time instances $\bm{t}^{(d)}$ for $d\in V$, $\bm{K}(\bm{t},\bm{t};\bm{\theta_g})\in\mathbb{R}^{N_v\times N_v}$ the following block diagonal kernel matrix:
\begin{equation}\label{eq:kernel_mat}
\bm{K}(\bm{t},\bm{t};\bm{\theta_g})=
\begin{bmatrix}
 \sum_{q=1}^{Q_{d_1}}k_{q,d_1}(\bm{t}^{(d_1)},\bm{t}^{(d_1)},\bm{\theta}_{d_1,q}) &  &  \\
 & \ddots &  \\
 &  & \sum_{q=1}^{Q_{d_{D_v}}}k_{q,d_{D_v}}(\bm{t}^{(d_{D_v})},\bm{t}^{(d_{D_v})},\bm{\theta}_{d_{D_v},q}) \\
\end{bmatrix} \ , 
\end{equation}
\noindent where $\{d_1,\ldots,d_{D_v}\}=V$, and $\bm{K_\epsilon}(\bm{\theta_g})$ the covariance matrix characterizing the noise process that may be corrupting the observations:
\begin{equation}\label{eq:kernel_mat_epsilon}
\bm{K_\epsilon}(\bm{\theta_g})=
\begin{bmatrix}
 \epsilon_{d_1}\mathbb{I}_{n^{(d_1)}} &  & \\
 & \ddots &  \\
 &  & \epsilon_{d_{D_v}}\mathbb{I}_{n^{(d_{D_v})}} \\
\end{bmatrix} \ , \{d_1,\ldots,d_{D_v}\}=V \ ,
\end{equation}
\noindent where we assume that we have a Gaussian noise of variance $\epsilon_d$ for the observable variable $d\in V$. $\bm{\theta_g}$ refers to the GP parameters and is given by:
\begin{equation}
\bm{\theta_g}= \{\bm{\theta}_{d,q},\epsilon_d, d\in V,q=1,\ldots,Q_d\} \ .
\end{equation}

The global parameter vector to be inferred is then given by:
\begin{equation}\label{eq:glob_par}
    \bm{\theta} = (\bm{\theta_f},\bm{\theta_g}) \ .
\end{equation}

%Using the Semi-Parameteric Latent Factor Model (SLFM) instead of the dimension-specific latent functions was also investigated. The SLFM is only valid if the time instances of the observations $\{t_i^{(d)}\}$, for $d\in V$, are the same. Therefore, it is computationally less expensive than the dimension-specific latent functions modeling thanks to the weights sharing across the different variables. However, the SLFM has a lower expressiveness than the dimension-specific latent functions modeling which was confirmed by the numerical experiments. Given the latter point and the necessity of developing a computational framework that has the ability to learn the dynamics from observations available at any time instances, we opted for the dimension-specific latent functions modeling. 

\subsection{Bayesian inference with Hamiltonian Monte Carlo}\label{sec:MCMC}

The Bayesian formalism provides a natural way to account for uncertainty, while also enabling the use of prior information for the unknown model parameters $\bm{\theta}$ (e.g. sparsity for $\bm{\theta_f}$ when using a dictionary as a representation of $f(\cdot,\cdot;\bm{\theta_f})$). More importantly, it enables the complete statistical characterization for all inferred parameters in our model. The latter, is encapsulated in the posterior distribution which can be factorized as:

\begin{equation}\label{equ:unnormalized_Bayes}
     p(\bm{\theta}| \mathcal{D}) = p(\bm{\theta_f},\bm{\theta_g} | \mathcal{D}) \varpropto p(\mathcal{D} | \bm{\theta_f},\bm{\theta_g} ) \ p(\bm{\theta_g}) \ p(\bm{\theta_f}),
\end{equation}
where $\mathcal{D}$ denotes the available data, and $p(\mathcal{D} | \bm{\theta_f},\bm{\theta_g})$ is a likelihood function that measures the discrepancy between the observed data and the model's predictions for a given set of parameters $(\bm{\theta_f},\bm{\theta_g})$. $p(\bm{\theta_g}) , p(\bm{\theta_f})$ are prior distributions that can help us encode any domain knowledge about the unknown model parameters $(\bm{\theta_f},\bm{\theta_g})$. Note that $\bm{\theta_g}$ accounts for the noise process that may be corrupting the observations. The posterior of equation (\ref{equ:unnormalized_Bayes}) is generally intractable, and approximate samples can be efficiently drawn using a state-of-the-art Hamiltonian Monte Carlo (HMC) \cite{neal2011mcmc} algorithm such as NUTS \cite{hoffman2014no} that automatically calibrates the step size of the Hamiltonian chain, and, consequently removes the need for hand-tuning it as opposed to relying on vanilla HMC \cite{yang2020bayesian}. 
% Such property generally provides very good convergence behavior and stabilizes the Bayesian inference process. Moreover, the log-determinant term in the GP likelihood is essentially a regularizer that favors finding the simplest possible model that best fits the observed data, which alleviate the burden of model selection.

Here, we employ a Bayesian approach corresponding to the following likelihood:
\begin{equation}
\label{eq:posterior}
\begin{aligned}
p(\mathcal{D} | \bm{\theta_f},\bm{\theta_g} ) = \mathcal{N}(\bm{z_V}(\bm{t};\bm{\theta_f}),\bm{K}(\bm{t},\bm{t};\bm{\theta_g})+\bm{K_\epsilon}(\bm{\theta_g})) \ ,
\end{aligned}
\end{equation}
\noindent where $\bm{z_V}(\bm{t})\in\mathbb{R}^{N_v}$ is the previously introduced concatenation of $\bm{z}_d$ at the time instances $\bm{t}^{(d)}$ for $d\in V$, and $\bm{z}(t;\bm{\theta_f})$ is the solution defined in (\ref{eq:NeuralODE}) and obtained with a differentiable ODE solver enabling gradient back-propagation \cite{chen2018neural}.
In addition to modeling temporal correlations in the observed state variables, the choice of a GP prior introduces an implicit regularization mechanism that favors finding the simplest possible model that best fits the observed data \cite{rasmussen2006gaussian}. This so-called Occam's razor principle \cite{MacKay1992} essentially enables automatic model selection without the need of resorting to empirical selection criteria for balancing the trade-off between model complexity and data-fit \cite{Pillonetto2014}.
% The latter is a general framework that was introduced in order to propagate  gradient information through classical numerical solvers for ordinary differential equations (ODEs) that blends classical adjoint methods \cite{pontryagin2018mathematical} with modern developments in automatic differentiation \cite{van2018automatic}. 

Given the observations forming the available data $\mathcal{D}$, we would like to learn the parameters $\bm{\theta}$ that best explain the observed dynamics as quantified by the discrepancy $L$ between the observed data and the model's predictions. A sufficient way to update the probability $p(\bm{\theta}|\mathcal{D})$ is through HMC \cite{neal2011mcmc}, however appropriate methods should be considered in order to effectively back-propagate gradients through the likelihood computation given the dependency on the dynamical system (\ref{equ:dynamics}), which requires a differentiable ODE solver. Such task can be carried out by defining the adjoint of the dynamical system as $\bm{a}(t) = \frac{\partial L}{\partial\bm{x}(t)}$. Then, the dynamical system describing the evolution of the adjoint can be derived as \cite{pontryagin2018mathematical, chen2018neural}:

\begin{equation}
    \frac{d\bm{a}(t)}{dt} = -\bm{a(t)}^{T}\frac{\partial f(\bm{x}(t), t, \bm{\theta})}{\partial \bm{x}}.
\end{equation}

Note that the adjoint $\bm{a}(t)$ can be computed by an additional call to the chosen ODE solver, and the target derivative $\frac{\partial L}{\partial\bm{\theta}}$ can be then computed as:

\begin{equation}\label{equ:adjoint}
    \frac{\partial L}{\partial\bm{\theta}} = - \int_{t_1}^{t_0}\bm{a(t)}^{T}\frac{\partial f(\bm{x}(t), t, \bm{\theta})}{\partial \bm{\theta}}dt,
\end{equation}
where $\frac{\partial f(\bm{x}(t), t, \bm{\theta})}{\partial \bm{\theta}}$ can be evaluated via automatic differentiation \cite{chen2018neural}.

% The purpose of the proposed method is to model the probabilistic conditional data pairs dependency $\{\bm{x}(t_i + \Delta t), \bm{x}(t_i)\}$. This would enable the parallel implementation of the model to significantly accelerate the training process. In other words, the only assumption is that the time difference between the initial data and the target is the same for all data pairs, denoted by $\Delta t$. 

Taken all together, the main advantages of the GP-NODE approach can be summarized in the following points:
\begin{itemize}
  \item The observed data does not need to be collected on a regular time grid.
  \item The data can be collected on different time grids for the different observed variables. In such scenario, the dynamical system (\ref{equ:dynamics}) needs to be solved for the collection of all time instances over which the data is available, and each variable is sub-sampled from the solution at the corresponding time instances for which the observations are available. Hence, there is no constrain to impose on $\bm{t}^d$, $d\in V$.
  \item The time-steps between different observations can be relatively large. For such large time steps, a classical numerical scheme can be used to integrate equation (\ref{equ:adjoint}) on a finer temporal discretization, with the latter being typically chosen according to the stability properties of the underlying ODE solver. 
\end{itemize}

%A systems identification task is now summarized as follows. Given the observations $\mathcal{D}$ detailed above one would like to learn the $\bm{\theta_f}$ that best parametrizes the underlying dynamics.

% discussion about neural ODE with proparagation given the IC

\subsection{Learning dynamics}
\label{sec:learning_dyanamics}

In this section we present two different problem settings that cover a broad range of practical applications and can be tackled  by the GP-NODE method. The first class consists of problems in which the model form of the underlying latent dynamics is completely unknown. In this setting, one can aim to distill a parsimonious and interpretable representation by constructing a comprehensive dictionary over all possible interactions and try to infer a predictive, yet minimal model form \cite{brunton2016discovering, rudy2017data, qin2019data, yang2020bayesian}. The second class of problems contains cases where a model form for the underlying dynamics is prescribed by domain knowledge, but a number of unknown parameters needs to be calibrated in order to accurately explain the observed data \cite{tartakovsky2018learning,yazdani2019systems}. The aforementioned problem classes are not mutually exclusive and can be combined in a hybrid manner to construct ``gray-box" models that are only partially informed by domain knowledge \cite{rackauckas2020universal, yang2020bayesian, rico1993continuous}.
As we will see in the following, the proposed workflow can seamlessly accommodate all aforementioned cases in a unified fashion, while remaining robust with respect to incomplete model parametrizations, as well as imperfections in the observed data. 
% Under this setup, our goal is to employ the algorithmic framework outlined in sections \ref{sec:diff_prog} and \ref{sec:MCMC} to perform probabilistic  inference over plausible sets of model parameters $\bm{\theta}$ that yield interpretable, parsimonious, and predictive representations. 

\subsubsection{Inferring an unknown model form via sparse dictionary learning}\label{sec:learning_dynamics_dict}
For the first problem setting mentioned above, only the data is available and there is no other prior knowledge. A data-driven approach is then used to learn the form of the dynamical system. Sparse identification has been investigated to identify the form of nonlinear dynamical systems \cite{brunton2016discovering}, where the right hand side of the dynamical system is approximated by a large dictionary, and the corresponding parameters for the different terms belonging to the dictionary are inferred.  
% Koopman operator methods are another alternative and can be used to learn the eigenvalues and modes of a corresponding linearized model \cite{schmid2010dynamic}. 
The proposed method considers a similar setup as in \cite{brunton2016discovering} where a dictionary is used to approximate the right hand side of the dynamical system. To this end, we parametrize the latent dynamics as:
\begin{equation}\label{equ:Discovery}
\begin{aligned}
    \frac{d\bm{x}}{dt} = A\varphi(\bm{x}),\\
\end{aligned}
\end{equation}
where $A\in\mathbb{R}^{D\times K}$ is a matrix of unknown dictionary coefficients, and $K$ the length of the dictionary $\varphi(\bm{x})\in\mathbb{R}^K$. $A$ contains the parameters that will be estimated and $\varphi(\bm{x})$ represents the prescribed dictionary of features (e.g., polynomials, Fourier modes, etc., and combinations thereof) \cite{brunton2016discovering, rudy2017data, champion2019data} that are used to represent the latent the dynamics.

Let $\bm{\theta_A}\in\mathbb{R}^{D\times K}$ be the vector containing the coefficients of $A$, such that $\bm{\theta_A}$ is a sub-vector of $\bm{\theta_f}$. In order to enable the discovery of interpretable and parsimonious representations for the underlying dynamics, we use the sparsity-promoting Finnish Horseshoe distribution as a prior \cite{piironen2017a} over the dictionary weights. The latter was originally developed for linear regression and can be detailed as follows:
\begin{equation}\label{equ:fh_prior}
\begin{gathered}
   \tau_0=\frac{m_0}{D\;K-m_0}\frac{1}{\sqrt{N_v}} \ , \\
   \tau\sim{\rm Half-}C(0,\tau_0) \ , \\
   c^2\sim{\rm Inv-}\mathcal{G}(\frac{\nu}{2},\frac{\nu}{2}s^2) \ , \\
   \lambda_m\sim{\rm Half-}C(0,1) \ , \ {\rm i.i.d} \ , \ m=1,\ldots,D\;K \ , \\
   \tilde{\lambda}_m=\frac{c\;\lambda_m}{c^2+\tau^2\;\lambda_m^2} \ , \\
   (\bm{\theta_A})_m\sim\mathcal{N}(0,\tau\;\tilde{\lambda}_m) \ ,
\end{gathered}
\end{equation}
\noindent where $\nu$ and $s$ are hyper-parameters which will be set equal to $1$ across all numerical examples thanks to the data normalization procedure we present in section \ref{sec:normalization}, and $m_0$ is the anticipated number of non-zero parameters. In standard applications of the Finnish Horseshoe in linear regression problems \cite{piironen2017a}, $m_0$ is simply fixed as a prior guess. In the GP-NODE framework, thanks to the incorporation of the physics knowledge through the ODE, $m_0$ will always be taken equal to its maximum possible value $D\times K-1$, independently of the true number of non-zero parameters. Calibrating the physics-based dynamical system with the observed data allows the GP-NODE framework to discover the actual number of non-zero parameters, which keeps the proposed method as general as possible, and without any hyperparameter tuning or manual incorporation of any anticipated results. The heavy-tailed Cauchy prior distribution for the local scales $\lambda_m$ allows the data to push those corresponding to non-zero parameters to large values as needed, which then pushes the corresponding parameters $(\bm{\theta_A})_m$ above the global scale $\tau$. Thanks to the parameterization of the latter, the data is able to refine the scale beyond the prior judgement of $\tau_0$. The scale $c$ is introduced as an another level to the prior hierarchy such that if we integrate it out of the distribution, it implies a marginal Student-t$(\nu,0,s)$ prior for each of the parameters $(\bm{\theta_A})_m$ which should be sufficiently far above the global scale $\tau$. Hence, the scale $c$ prevents the parameters $(\bm{\theta_A})_m$ from surpassing the global scale $\tau$ and thus being un-regularized, which would result in posteriors that diffuse to significantly large values. Such behavior would yield to a poor identification of the likelihood to propagate to the posterior and to inaccurate inference results.

The advantage of the Horseshoe and Finnish Horseshoe priors over the Laplace prior employed in \cite{yang2020bayesian} stems from the difference in the ``shrinkage effects" \cite{Carvalho2009,piironen2017a,Piironen2017b}, which can be interpreted as the amount of weight that the posterior mean for the inferred parameters $\bm{\theta_A}$ places on the value of $0$ once the data has been observed. In this regard, the most important characteristic of the  Horseshoe and Finnish Horseshoe priors is the distinguished separation between the global and local shrinkage effects thanks to the use of the global scale $\tau$ and local scales $\lambda_m$: the global scale tries to estimate the overall sparsity level, while the local scales permit flagging the non-zero elements of the inferred parameters $\bm{\theta_A}$. The heavy-tailed Cauchy prior distributions used for the local scales $\lambda_m$ play a crucial role in this process by allowing the estimated non-zero elements of $\bm{\theta_A}$ to escape the strong pull towards the value of $0$ applied by the global scale $\tau$. Hence, the Horseshoe and Finnish Horseshoe priors have the ability to shrink globally thanks to the global scale $\tau$, and yet act locally thanks to the local scales $\lambda_m$, which is not possible under the Laplace prior whose shrinkage effect imposes a compromise between shrinking zero parameters and flagging non-zero ones. Such characteristic often results to overestimation of sparsity and under-estimation of the larger non-zero parameters \cite{Carvalho2009}.

\subsubsection{Incorporating domain knowledge}\label{sec:learning_dynamics_knowledge}
The second problem setting involves parameter discovery, which corresponds to the setting where prior knowledge of the underlying model form is available. Such physics-informed knowledge can sometimes be critical to learn certain important parameters related to physical phenomenon such as limit cycles in chemical system \cite{schnakenberg1979simple}, shock waves in fluid dynamics \cite{raissi2018hidden}, evolution of biological systems \cite{yazdani2019systems}, development of Darcy flow \cite{tartakovsky2018learning}, etc. In this context, the exact form of the right hand side of the dynamical system as given in equation (\ref{equ:dynamics}) is available. In other words, the form of $f(\cdot,\cdot;\bm{\theta_f})$ is assumed to be known.

\subsection{Generating forecasts with quantified uncertainty}\label{sec:predictive_inference}

The Hamiltonian Monte Carlo procedure described in section \ref{sec:MCMC} produces a set of samples that concentrate in regions of high-probability in the posterior distribution $p(\bm{\theta} |\mathcal{D})$. Approximating this distribution is central to our workflow as it enables the generation of future forecasts $\bm{x}^*(t^*)$ at time instances $\bm{t}^*$ of size $N^*$ with quantified uncertainty via computing the predictive posterior distribution: 

\begin{equation}\label{equ:predictive_distribution}
\begin{aligned}
    p(\bm{x}^*(\bm{t}^*)|\mathcal{D}, \bm{t}^*) = \int p(\bm{x}^*(\bm{t}^*)|\bm{\theta}, t^*)p(\bm{\theta}|\mathcal{D} )d\bm{\theta}.
%    h_{\bm{\theta}}(\bm{x}_0, t), 
\end{aligned}
\end{equation}
This predictive distribution provides a complete statistical characterization for the forecasted states by encapsulating epistemic uncertainty in the inferred dynamics, as well as accounting for the fact that the model was trained on a finite set of noisy observations.
This allows us to generate GP-NODE-based plausible realizations of $\bm{x}^*(\bm{t}^*,\bm{\theta})$ by sampling from the predictive posterior distribution as:
\begin{equation}
\begin{gathered}
    \frac{d \bm{h}}{dt} = f(\bm{h}, t; \bm{\theta_f}) \ , \\ \\
    \bm{x}^*_{d'}(\bm{t}^*,\bm{\theta})=\bm{h}_{d'}(\bm{t}^*,\bm{\theta_f}) \ , d'\notin V \ ,\\ \\
    \bm{x}^*_{d}(\bm{t}^*,\bm{\theta}) \sim \mathcal{N}\big(\bm{\mu}^*_d(\bm{t}^*,\bm{\theta}), \bm{K}^*(\bm{t}^*,\bm{\theta_g})\big) \ , d\in V \ , \\ \\
    \bm{\mu}^*_d(\bm{t}^*,\bm{\theta}) = \bm{h}_{d}(\bm{t}^*,\bm{\theta_f}) + \bm{K}(\bm{t}^*,\bm{t};\bm{\theta_g})\; \big(\bm{K}(\bm{t},\bm{t};\bm{\theta_g})+\bm{K_\epsilon}(\bm{\theta_g})\big)^{-1}\;
    \big(\bm{x}_d(\bm{t}^{(d)})-\bm{h}_d(\bm{t}^{(d)},\bm{\theta_f})\big) \ ,
\end{gathered}
\end{equation}
\noindent where $\bm{\theta_f} \sim p(\bm{\theta_f}|\mathcal{D})$ and $\bm{\theta_g} \sim p(\bm{\theta_g}|\mathcal{D})$ are approximate samples from $p(\bm{\theta_f}, \bm{\theta_g}|\mathcal{D})$ computed during model training via HMC sampling; $\bm{K}(\bm{t},\bm{t};\bm{\theta_g})$ and $\bm{K_\epsilon}(\bm{\theta_g})$ are defined in equations (\ref{eq:kernel_mat}) and ($\ref{eq:kernel_mat_epsilon}$), respectively, and

\begin{equation}
\begin{gathered}
\bm{K}^*(\bm{t}^*,\bm{\theta_g})= \bm{K}(\bm{t}^*,\bm{t}^*;\bm{\theta_g})-\bm{K}(\bm{t}^*,\bm{t};\bm{\theta_g})\;(\bm{K}(\bm{t},\bm{t};\bm{\theta_g})+\bm{K_\epsilon}(\bm{\theta_g}))^{-1}\;\bm{K}(\bm{t}^*,\bm{t};\bm{\theta_g})^T \ , \\ \\
\bm{K}(\bm{t}^*,\bm{t}^*;\bm{\theta_g})=
\begin{bmatrix}
 \sum_{q=1}^{Q_{d_1}}k_{q,d_1}(\bm{t}^*,\bm{t}^*,\bm{\theta}_{d_1,q}) &  &  \\
 & \ddots &  \\
 &  & \sum_{q=1}^{Q_{d_{D_v}}}k_{q,d_{D_v}}(\bm{t}^*,\bm{t}^*,\bm{\theta}_{d_{D_v},q}) \\
\end{bmatrix} \ , \\ \\
\bm{K}(\bm{t}^*,\bm{t};\bm{\theta_g})=
\begin{bmatrix}
 \sum_{q=1}^{Q_{d_1}}k_{q,d_1}(\bm{t}^*,\bm{t}^{(d_1)},\bm{\theta}_{d_1,q}) &  &  \\
 & \ddots &  \\
 &  &  & \sum_{q=1}^{Q_{d_{D_v}}}k_{q,d_{D_v}}(\bm{t}^*,\bm{t}^{(d_{D_v})},\bm{\theta}_{d_{D_v},q}) \\
\end{bmatrix} \ , 
\end{gathered}
\end{equation}
\noindent where $\{d_1,\ldots,d_{D_v}\}=V$. $\bm{h}(\bm{t}^*,\bm{\theta_f})$ denotes any numerical integrator that predicts the system's state at time instances $\bm{t}^*$.

Finally, it is straightforward to utilize the posterior samples of $\bm{\theta}\sim p(\bm{\theta}|\mathcal{D})$ to approximate the first- and second-order statistics of the predicted states $\bm{x}^*(\bm{t}^*)$ as
\begin{align}
    \label{eq:statistics}
    \hat{\mu}_{\bm{x}^*}(\bm{t}^*) & = \int \bm{x}^*_{\bm{\theta}}(\bm{t}^*) p(\bm{\theta}|\mathcal{D})d\bm{\theta} \approx \frac{1}{N_s}\sum\limits_{i=1}^{N_s} \bm{x}^*_{\bm{\theta}_i}(\bm{t}^*),\\
    \hat{\sigma}^{2}_{\bm{x}^*}(\bm{t}^*) & = \int [ \bm{x}^*_{\bm{\theta}}(\bm{t}^*) - \hat{\mu}_{\bm{x}^*}(\bm{t}^*) ]^2 p(\bm{\theta}|\mathcal{D})d\bm{\theta} \approx \frac{1}{N_s}\sum\limits_{i=1}^{N_s} [ \bm{x}^*_{\bm{\theta}_i}(\bm{t}^*) - \hat{\mu}_{\bm{x}^*}(\bm{t}^*) ]^2,
\end{align}
\noindent where $N_s$ denotes the number of samples drawn via  Hamiltonian Monte Carlo to simulate the posterior, i.e., $\bm{\theta}_i\sim p(\bm{\theta}|\mathcal{D})$, $i=1,\dots,N_s$. Note that higher-order moments are also readily computable in a similar manner.

\subsection{Model initialization, priors and data pre-processing}\label{sec:normalization}
To promote robustness and stability of the proposed GP-NODE framework across different numerical examples, the training data is appropriately normalized in order to prevent gradient pathologies during back-propagation \cite{glorot2010understanding}. Hence, the observable variables are normalized as follows:
\begin{equation}
\label{eq:normalization}
    \tilde{\bm{x}}_d = \frac{\bm{x}_d}{\max\limits_{i=1,\ldots,n^{(d)}}|\bm{x}_d(t_i^{(d)})|} \ , \ d\in V \ .
\end{equation}
The time instances, which are used as input to the Gaussian processes, are also normalized as follows:
\begin{equation}
\label{eq:normalization_time}
    \tilde{t}_i^{(d)} = \frac{t_i^{(d)}}{\max\limits_{i=1,\ldots,n^{(d)} \ , \ d\in V}t_i^{(d)}} \ , \ d\in V \ .
\end{equation}
For all the problems considered in section \ref{sec:Results}, modeling each of the observable variables using independent Gaussian process priors is sufficient to obtain satisfactory results. Hence, we have chosen $Q_d=1$ for $d\in V$ (see equation (\ref{eq:kernel_mat})). Temporal correlations in the observed data are captured via employing an exponentiated quadratic kernel which implicitly introduces a smoothness assumption for the latent dynamics, such that:
\begin{equation}\label{eq:kernel}
    k_{q,d}(t,t',\bm{\theta}_{d,q})=w_{d,q}e^{-\frac{(t-t')^2}{l_{d,q}^2}} \ , d\in V \ , q=1 \ ,
\end{equation}
\noindent where $\bm{\theta}_{d,q}\equiv(w_{d,q},l_{d,q})$. Note that other kernel choices can be readily incorporated to reflect more appropriate prior assumptions for a given problem setting.

Data normalization enables the use of the following unified priors across the all numerical examples
\begin{equation}\label{eq:priors}
\begin{aligned}
    w_{d,q}\sim{\rm Lognormal}(0,1) \ , \ q=1,\ldots,Q_d \ , \ d\in V \ , \\
    l_{d,q}\sim{\rm Gamma}(\alpha=1,\beta=1/2) \ , \ q=1,\ldots,Q_d \ , \ d\in V \ , \\
    \epsilon_d\sim{\rm Lognormal}(0,1) \ , \ q=1,\ldots,Q_d \ , \ d\in V \ .
\end{aligned}
\end{equation}

The prior distributions considered for the parameters $\bm{\theta_f}$ depend on the class of problems detailed in section \ref{sec:learning_dyanamics}. For problems for which the undelrying model form is known, the prior distributions of $\bm{\theta_f}$ are defined based on the available domain knowledge. For problem settings with an unknown model form, the Finnish horseshoe distribution, detailed in section \ref{sec:learning_dynamics_dict}, is considered as a prior for the dictionary parameters, while the priors of any remaining parameters are defined based on the available domain knowledge. Finally, for all the numerical examples presented in section \ref{sec:Results}, the NUTS sampler \cite{hoffman2014no} is used with $4,000$ warmup (burn-in) steps, followed by $8,000$ HMC iterations with a target acceptance probability equal to $0.85$.

\section{Results}
\label{sec:Results}

In this section, we present a comprehensive collection of numerical studies to illustrate the key features of the GP-NODE methodology, and place into context against existing state-of-the-art methods for systems identification such as the SINDy framework of Brunton {\em et al.} \cite{brunton2016discovering}, and of the Nonparametric ODE Model method (npODE) of Heihonen {\it et al.} \cite{Heinonen2018} which has been shown to outperform the Gaussian Process Dynamical Model approach (GPDM) \cite{Wang2006} and the Variational Gaussian Process Dynamical System method (VGPDS) \cite{Damianou2011}. Specifically, we expand on three benchmark problems that cover the problem settings discussed in section \ref{sec:learning_dyanamics}. The algorithmic settings used across all cases follow the discussion provided in \ref{sec:normalization}. Our implementation leverages GPU hardware and automatic differentiation via JAX \cite{jax2018github} and the NumPyro library for probabilistic programming \cite{phan2019composable,bingham2018pyro}. All code and data presented in this section will be made publicly available at \url{https://github.com/PredictiveIntelligenceLab/GP-NODEs}.

% the two contexts introduced previously (the dictionary learning context and the \textcolor{red}{domain knowledge enabled learning}). (1) For the dictionary learning context, a two-dimensional damped oscillator, a two-dimensional damped pendulum and a Lorenz system are considered to illustrate the usefulness of the proposed Bayesian \textcolor{red}{differential programming}. (2) For the \textcolor{red}{domain knowledge enabled learning} context, the Lotka–Volterra equations that exhibit limit cycle behavior is considered, along with a real world biological system: the Yeast Glycolysis system.

\subsection{Dictionary learning for a predator-prey system with with time gaps, different time grids and unknown initial conditions}\label{sec:LV}
This case study is designed to illustrate the capability of the GP-NODE framework to accommodate noisy data, observations with a time gap and at different time instances, and unknown initial conditions; a common practical setting that cannot be effectively handled by popular systems identification methods \cite{brunton2016discovering,rudy2017data,lusch2018deep,raissi2018multistep, qin2019data}. To this end, let us consider a classical prey-predator system described by the Lotka–Volterra equations:
\begin{equation}\label{eq:LV}
\begin{aligned}
    \frac{d x_1}{dt} = \alpha x_1 + \beta x_1 x_2 \ ,\\
    \frac{d x_2}{dt} = \delta x_1 x_2 + \gamma x_2 \ ,
\end{aligned}
\end{equation}
which is known to exhibit a stable limit cycle behavior for $\alpha = 1.0, \beta = -0.1, \gamma = -1.5$, and $\delta = 0.75$. Our goal here is to recover this dynamical system directly from data using a polynomial dictionary parametrization taking the form:

\begin{equation}
\begin{bmatrix}
    \frac{dx_1}{dt}\\
    \frac{dx_2}{dt}
\end{bmatrix}
= A\varphi(\bm{x}) = 
\begin{bmatrix}
  \textcolor{red}{a_{11}} & a_{12} & \textcolor{red}{a_{13}} & a_{14} & a_{15} & a_{16} & a_{17} \\
  a_{21} & \textcolor{red}{a_{22}} & \textcolor{red}{a_{23}} & a_{24} & a_{25} & a_{26} & a_{27}
\end{bmatrix}
\begin{bmatrix}
   x_1 \\ x_2 \\ x_1 x_2 \\ x_1^2 \\ x_2^2 \\ x_1^3 \\ x_2^3
\end{bmatrix},
\end{equation}
where the $a_{ij}$'s are unknown scalar coefficients to be  estimated. The active non-zero parameters are highlighted with red color for clarity. To generate the training data, we simulate the exact dynamics of equation (\ref{eq:LV}) in the time interval $t\in [0, 16.5]$ with $\alpha = 1.0, \beta = -0.1, \gamma = -1.5, \delta = 0.75$ and the initial condition: $(x_1, x_2) = (5, 5)$. The training data-set $\mathcal{D}$ is considered such that the simulated data is perturbed by $10\%$ white noise. Both variables for this example are taken as observable ($V=\{1,2\}$) but their corresponding observations are sampled at different time instances as follows:
\begin{equation}\label{eq:LV_t}
    \begin{aligned}
        \bm{t}^{(1)}=(1.5,1.8,2.1,\ldots,15.9,16.2,16.5) \ , \\
        \bm{t}^{(2)}=(0,1.65,1.95,2.25,\ldots,16.05,16.35,16.5) \ .
    \end{aligned}
\end{equation}
\noindent 
As such, we have $51$ observations for variable $x_1$ with a time gap of $1.5$, a time step equal to $0.3$, and an unknown initial condition. We also have $52$ observations for variable $x_2$ with a time gap of $1.65$ and a time step equal to $0.3$. For this example, the dynamics parameters vector is given by $\bm{\theta_f}=(\bm{\theta_A},x_{1,0})$, where $\bm{\theta_A}$ is the vector containing the coefficients of $A$ and $x_{1,0}$ is the inferred initial condition. Given this noisy, sparse and irregular training data, our goal is to demonstrate the performance of the GP-NODE framework in identifying the unknown model parameters $\bm{\theta} = (\bm{\theta_f},\bm{\theta_g})$ (see equation (\ref{eq:glob_par})) by inferring their posterior distributions. The Finnish Horseshoe distribution is used as prior for the parameters $\bm{\theta_A}$ as detailed in section \ref{sec:learning_dynamics_dict}, while the prior distribution of $\bm{\theta_g}$ is discussed in section \ref{sec:normalization}. Finally, the prior distribution of $x_{1,0}$ is taken as the uniform distribution over $[4,6]$.

\begin{figure}
\centering
\includegraphics[width=\textwidth]{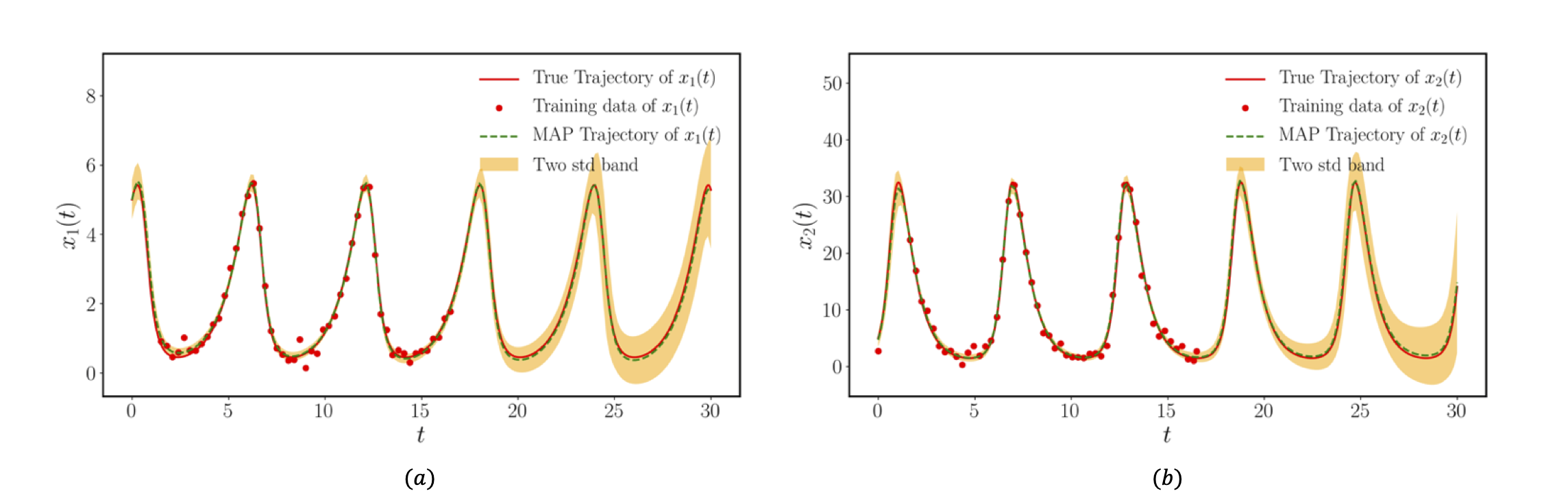}
\caption{{\em Parameter inference in a predator-prey system:} (a) Learned dynamics versus the true dynamics and the training data of $x_1(t)$. (b) Learned dynamics versus the true dynamics and the training data of $x_2(t)$.}
\label{fig:Diff_t_t_gap_B_m0_Mm1_2_chains_LV_x0_GP}
\end{figure}

\begin{figure}
\centering
\includegraphics[width=\textwidth]{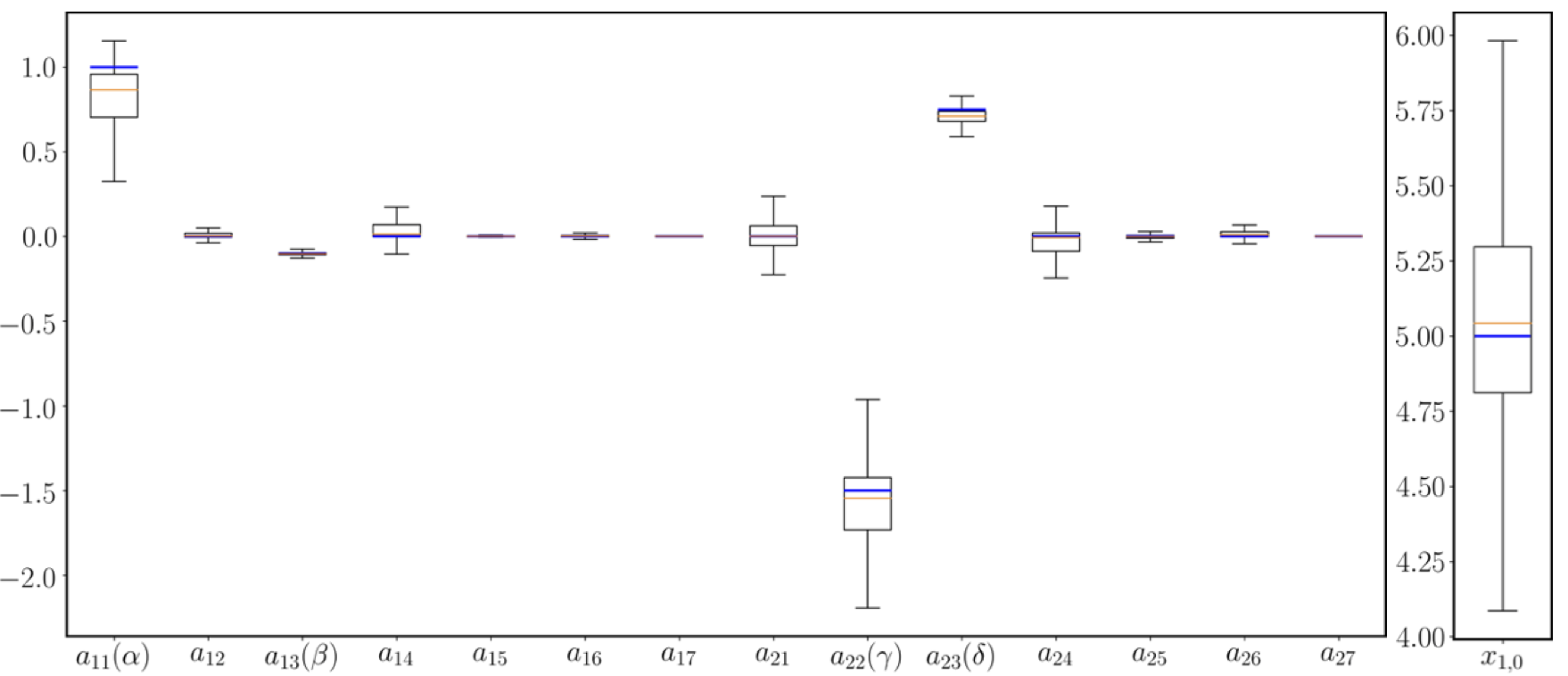}
\caption{{\em Predator-prey system dynamics:} Uncertainty estimation of the inferred model parameters obtained using the GP-NODE method. Estimates for the minimum, maximum, median, first quantile and third quantile are provided, while the true parameter values are highlighted in blue.}
\label{fig:B_m0_Mm1_2_chains_box_no_crop}
\end{figure}

Our numerical results are presented in figure \ref{fig:Diff_t_t_gap_B_m0_Mm1_2_chains_LV_x0_GP} showing the training data and the estimated trajectories based on the inferred parameters. Uncertainty estimates for the inferred parameters can also be deducted from the computed posterior distribution $p(\bm{\theta_f}|\mathcal{D})$, as presented in the box plots of figure \ref{fig:B_m0_Mm1_2_chains_box_no_crop} where the minimum, maximum, median, first quantile and third quantile obtained from the HMC simulations for each parameter are presented. %Finally, notice that all true values fall between the predicted quantiles.

It is evident that the GP-NODE approach is: (i) able to provide an accurate estimation for the unknown model parameters, (ii) yield an estimator with a predicted trajectory that closely matches the exact system's dynamics, (iii) return a posterior distribution over plausible models that captures both the epistemic uncertainty of the assumed parametrization and the uncertainty induced by training on a finite amount of noisy training data, and (iv) propagate this uncertainty through the system's dynamics to characterize variability in the predicted future states.

\begin{figure}
\centering
\includegraphics[width=\textwidth]{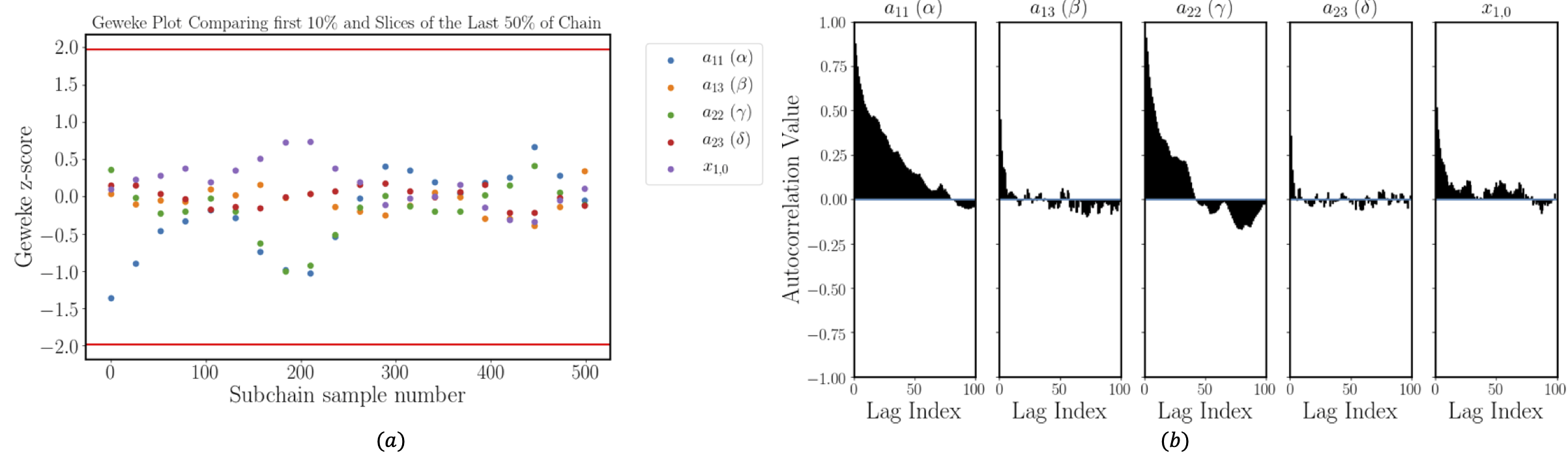}
\caption{\textcolor{black}{{\em HMC convergence diagnostics for the predator-prey system parameters $a_{11}, a_{13}, a_{22}, a_{23}$, $x_{1,0}$}: (a) The Geweke diagnostic test based on the chain of $1000$ samples: we compare the mean of the first  10\% of the chain of $1000$ samples to the mean of $20$ sub-samples from the last 50\% of the chain. (b) Auto-correlation as function of the lag.}}
\label{fig:LV_MCMC_diagnostics}
\end{figure}

To verify the statistical convergence of the HMC chains, we have employed the Geweke diagnostic and Gelman Rubin tests \cite{geweke1993bayesian,gelman2013bayesian}. The Gelman Rubin tests have provided the following values for the non-zero parameters ($a_{11}: 1.000, a_{13}: 1.604, a_{22}: 1.191, a_{23}: 1.012$) and the inferred initial condition ($x_{1,0}: 1.159$), which are all close to the optimal value of $1$, as expected. In order to perform the Geweke diagnostic, we have ``thinned" the $8,000$ HMC draws by randomly sampling one data point every eight consecutive steps of the chain. In order to implement the Geweke diagnostic, we considered the first 10\% of the chain of the $1000$ samples and compared their mean with the mean of $20$ sub-samples from the last 50\% of the chain. As shown in plot (a) of figure \ref{fig:LV_MCMC_diagnostics}, the Geweke z-scores for the four non-zero parameters are well between $-2$ and $2$. The results of auto-correlation are given in plot (b) of figure \ref{fig:LV_MCMC_diagnostics}. The Geweke diagnostic, the Gelman Rubin tests results and the auto-correlation values demonstrate a ``healthy" mixing behavior of the NUTS sampler \cite{hoffman2014no}.

%Another interesting observation here is that the Markov Chain Monte Carlo sampler is very efficient in identifying the importance/sensitivity of each inferred parameter in the model. For instance, less important parameters have the highest uncertainty, as observed in the posterior density plots shown in figures \ref{fig:LV_noise_free_data}(b) and \ref{fig:LV_noisy_data}(b). Specifically, notice how the posterior distribution of $\alpha$ and $\gamma$ has a considerably larger standard deviation than the other parameters, implying that the evolution of this dynamical system is less sensitive with respect to these parameters.

%This example is provided with the normalization trick to demonstrate the power of the proposed method. Noting that the largest standard deviation of the three dimensions $[x_1, x_2]$ as $\sigma_{\textrm{normal}}$. The normalized dynamical system can be obtained as:
%\begin{equation}
%\begin{aligned}
%    \frac{d\tilde{x}_1}{dt} = \alpha \tilde{x}_1 - \beta \sigma_{\textrm{normal}}\tilde{x}_1\tilde{x}_2,\\
%    \frac{d\tilde{x}_2}{dt} = \delta \sigma_{\textrm{normal}}\tilde{x}_1\tilde{x}_1 - \gamma \tilde{x}_2.
%\end{aligned}
%\end{equation}

\begin{figure}
\centering
\includegraphics[width=\textwidth]{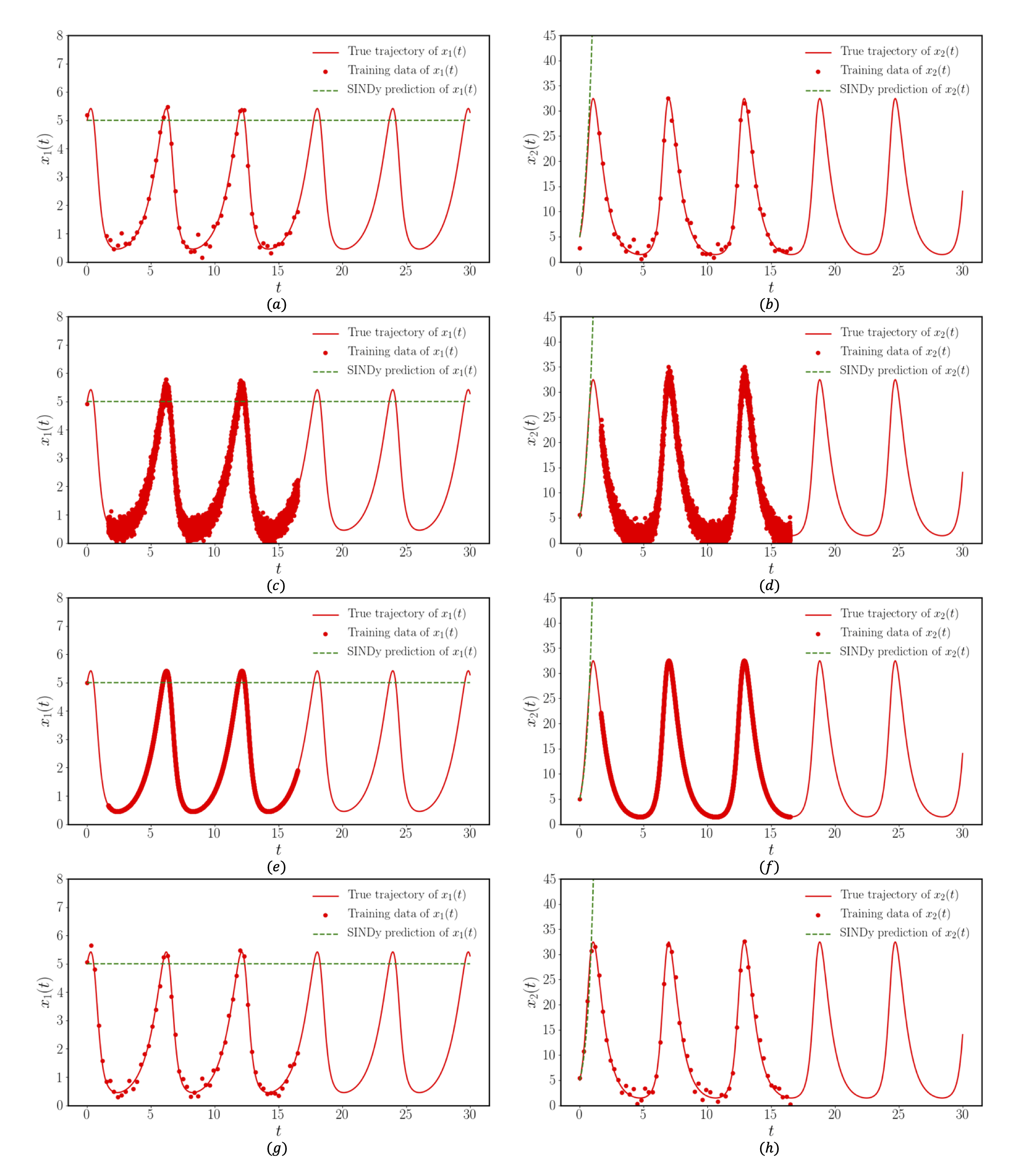}
\caption{{\em Parameter inference in a predator-prey system:} (a), (b) SINDy's predictions for $x_1(t)$ and $x_2(t)$ respectively versus the true dynamics and the training data for case 1. (c), (d) SINDy's predictions for $x_1(t)$ and $x_2(t)$ respectively versus the true dynamics and the training data for case 2. (e), (f) SINDy's predictions for $x_1(t)$ and $x_2(t)$ respectively versus the true dynamics and the training data for case 3. (g), (h) SINDy's predictions for $x_1(t)$ and $x_2(t)$ respectively versus the true dynamics and the training data for case 4. }
\label{fig:sindy_fail}
\end{figure}

Finally, we show the merits of the GP-NODE method by comparing it against the SINDy algorithm of Brunton {\em et. al.} \cite{brunton2016discovering} for the benchmark presented above with the following simplifying modifications. First, the deterministic SINDy approach cannot accommodate inference over unknown initial conditions. Hence, here we will assume the latter as known. Moreover, SINDy requires that the observations provided for the different variables are sampled at the same time instances. Hence $x_1$ and $x_2$ observations are provided at the following time instances:
\begin{equation}\label{eq:LV_t_SINDy}
        \bm{t}^{(1)}=\bm{t}^{(2)}=(0,1.5,1.8,2.1,\ldots,15.9,16.2,16.5) \ .
\end{equation}
To this end, we have $52$ observations for each of the variables with a time gap of $1.5$ and a time step equal to $0.3$. Under this problem setup (case 1), the SINDy algorithm fails and cannot infer the parameters appropriately as shown in the first row of table \ref{tab:PP_SINDy_fail} and in plots (a) and (b) of figure \ref{fig:sindy_fail}.

We kept decreasing the time step but never obtained the right inference with SINDy. The smallest time step considered was $dt=0.00234375$ such that the data contains the initial condition plus $6,402$ observations between $t=1.5$ and $t=16.5$ for each of the variables. The results obtained with SINDy under such setup (case 2) are given in the second row of  table \ref{tab:PP_SINDy_fail} and in plots (c) and (d) of figure \ref{fig:sindy_fail}. We further simplified the problem by considering the previous setup (case 2), but with noise free data (case 3) and we still obtained erroneous predictions with SINDy, as shown in the third row of  table \ref{tab:PP_SINDy_fail}, and in plots (e) and (f) of figure \ref{fig:sindy_fail}. As a consequence, since considering noise free data and an extremely small time step did not help us obtain good results with the SINDy algorithm, it is clear that the latter is not able to handle the presence of a temporal gap in the observations for the predator-prey system considered.   

\begin{table}
\centering
\begin{tabular}{|c|c|c|c|c|c|c|c|c|c|c|c|c|c|c|c|}
\hline
 & $\textcolor{red}{a_{11}}$ & $a_{12}$ & $\textcolor{red}{a_{13}}$ & $a_{14}$ & $a_{15}$ & $a_{16}$ & $a_{17}$ & $a_{21}$ & $\textcolor{red}{a_{22}}$ & $\textcolor{red}{a_{23}}$ & $a_{24}$ & $a_{25}$ & $a_{26}$ & $a_{27}$ \\
\hline
case 1 & \textcolor{red}{0} & 0 & \textcolor{red}{0} & 0 & 0 & 0 & 0 & -1.619 & \textcolor{red}{-1.039} & \textcolor{red}{0.501} & 0.779 & 0 & 0 & 0\\
\hline
case 2 & \textcolor{red}{0} & 0 & \textcolor{red}{0} & 0 & 0 & 0 & 0 & -11.803 & \textcolor{red}{-0.977} & \textcolor{red}{0.511} & 7.057 & 0 & -0.882 & 0\\
\hline
case 3 & \textcolor{red}{-1.499} & 0 & \textcolor{red}{0.750} & 0 & 0 & 0 & 0 & 0 & \textcolor{red}{0} & \textcolor{red}{0} & 0 & 0 & -0 & 0\\
\hline
case 4 & \textcolor{red}{0} & 0 & \textcolor{red}{0} & 0 & 0 & 0 & 0 & -0.952 & \textcolor{red}{-1.330} & \textcolor{red}{0.677} & 0.194 & 0 & 0 & 0\\
\hline
\end{tabular}
\caption{{\em Dictionary learning for a predator-prey system:} Point estimates for the dictionary coefficients obtained by the  SINDy algorithm \cite{brunton2016discovering} for cases $1,\ldots,4$. The model's active non-zero parameters are highlighted with red color for clarity.}
\label{tab:PP_SINDy_fail}
\end{table}

Next, we extend our comparison by removing the time gap in the observations. At first, we keep the same time step ($dt = 0.3$) as above, and generate training data for variables $x_1$ and $x_2$ at the following time instances:
\begin{equation}\label{eq:LV_t_SINDy_2}
        \bm{t}^{(1)}=\bm{t}^{(2)}=(0,0.3,0.6,0.9,\ldots,15.9, 16.2,16.5) \ .
\end{equation}
Under this problem setup (case 4), the SINDy algorithm still fails because of the sparsity of the observations and cannot infer the parameters appropriately as shown in the fourth row of table \ref{tab:PP_SINDy_fail}, and in plots (g) and (h) of figure \ref{fig:sindy_fail}.

\begin{figure}
\centering
\includegraphics[width=\textwidth]{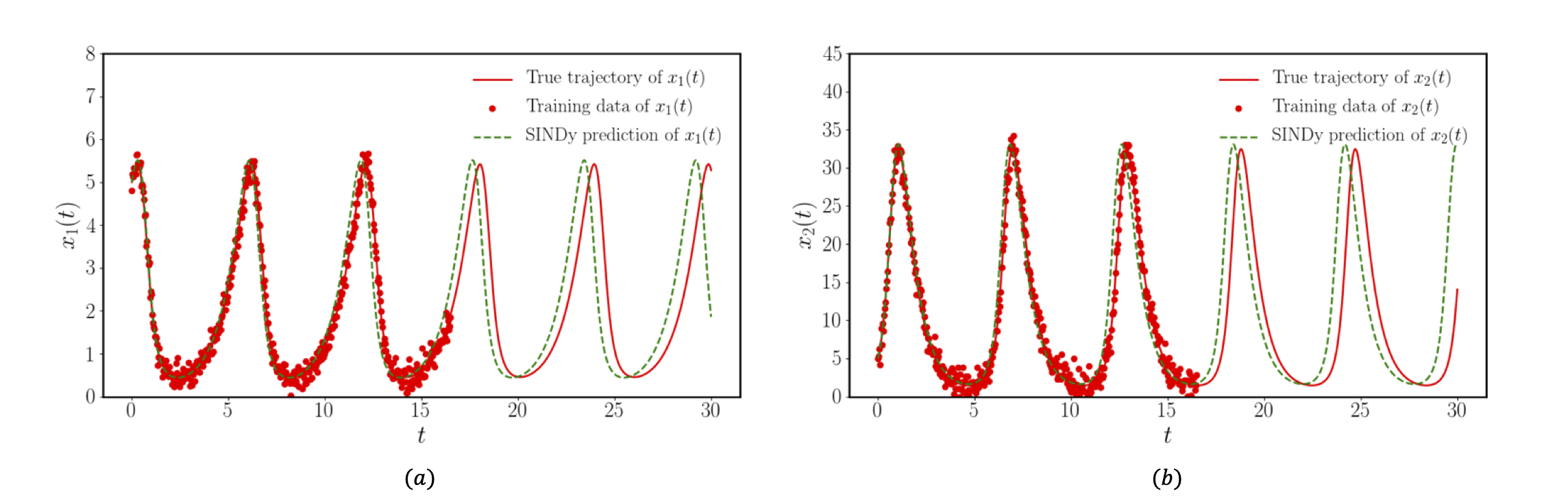}
\caption{{\em Parameter inference in a predator-prey system:} (a) SINDy's prediction versus the true dynamics and the training data of $x_1(t)$ for case 5. (b) SINDy's prediction versus the true dynamics and the training data of $x_2(t)$ for case 5.}
\label{fig:sindy_work}
\end{figure}

Next, we proceed by decreasing the time step until SINDy is able to provide a reasonable identification of the predator-pray dynamics. The required time step was $dt=0.0375$ such that the data does not contain any time gap in the observations, corresponding to $441$ equally spaced samples between $t=0$ and $t=16.5$ for each of the state variables. The results obtained with SINDy under this setup (case 5) are given in the table \ref{tab:PP_SINDy_work}, and in panels (a) and (b) of figure \ref{fig:sindy_work}. Although the inferred values for the parameters are close to the exact ones, the small difference between them results in predicted trajectories that deviate from the true ones, with the deviation being more significant as we move further in time. Compared to the predictions obtained with the GP-NODE methodology in figure \ref{fig:Diff_t_t_gap_B_m0_Mm1_2_chains_LV_x0_GP}, the differences between the MAP trajectories obtained with the GP-NODE and the true trajectories are clearly smaller than the differences obtained with the SINDy algorithm, although we considerably simplified the problem setup in order to favor the performance of SINDy by removing the time lag in the observations, not inferring any initial condition, assuming that both variables have observations at the same time instances, and reducing the time step of the trajectories used for training. Moreover, the GP-NODE method is a Bayesian approach and provides uncertainty estimates for the inferred parameters and also for future forecasts of the variable trajectories, which is obviously missing in the deterministic SINDy framework.

\begin{table}
\centering
\begin{tabular}{|c|c|c|c|c|c|c|c|c|c|c|c|c|c|c|c|}
\hline
$\textcolor{red}{a_{11}}$ & $a_{12}$ & $\textcolor{red}{a_{13}}$ & $a_{14}$ & $a_{15}$ & $a_{16}$ & $a_{17}$ & $a_{21}$ & $\textcolor{red}{a_{22}}$ & $\textcolor{red}{a_{23}}$ & $a_{24}$ & $a_{25}$ & $a_{26}$ & $a_{27}$ \\
\hline
\textcolor{red}{1.066} & 0 & \textcolor{red}{-0.102} & 0 & 0 & 0 & 0 & 0 & \textcolor{red}{-1.465} & \textcolor{red}{0.730} & 0 & 0 & 0 & 0\\
\hline
\end{tabular}
\caption{{\em Dictionary learning for a predator-prey system:} Point estimates for the dictionary coefficients obtained by the  SINDy algorithm \cite{brunton2016discovering} for case $5$. The model's active non-zero parameters are highlighted with red color for clarity.}
\label{tab:PP_SINDy_work}
\end{table}

\subsection{Bayesian calibration of a Yeast Glycolysis model}\label{sec:Glycolysis}
This example aims to illustrate the performance of the proposed GP-NODE framework applied to a more realistic setting involving biological systems. To this end, we consider a yeast glycolysis process which can be described by a 7-dimensional dynamical system \cite{ruoff2003temperature, yazdani2019systems} of the form:
\begin{equation}\label{eq:Glycolysis_system}
\begin{aligned}
    \frac{d S_1}{dt} &= J_0 - v_1, \\ 
    \frac{d S_2}{dt} &= 2v_1 - v_2 - v_6, \\
    \frac{d S_3}{dt} &= v2 - v_3, \\ 
    \frac{d S_4}{dt} &= v_3 - v_4 - J, \\
    \frac{d N_2}{dt} &= v_2 - v_4 - v_6, \\ 
    \frac{d A_3}{dt} &= - 2v_1 + 2v_3 - v_5, \\
    \frac{d S_{4}^{ex}}{dt} &= \phi J - v_7,
\end{aligned}
\end{equation}
where the terms $v_1, v_2, v_3, v_4, v_5, v_6, v_7, N_1$ and $A_2$ on the right hand side are defined as:
\begin{equation}\label{eq:Glycolysis_access}
\begin{aligned}
    v_1 &= k_1 S_1 A_3 [1 + (\frac{A_3}{K_I})^q]^{-1}, \\ 
    v_2 &= k_2 S_2 N_1, \\
    v_3 &= k_3 S_3 A_2, \\
    v_4 &= k_4 S_4 N_2, \\
    v_5 &= k_5 A_3, \\
    v_6 &= k_6 S_2 N_2, \\
    v_7 &= k S_{4}^{ex}, \\
    N_1 + N_2 &= N, \\
    A_2 + A_3 &= A.
\end{aligned}
\end{equation}
Here we assume that this model form is known from existing domain knowledge. A training data-set is generated from a single simulated trajectory of the system with an initial condition: $(S_1, S_2, S_3, S_4, N_2, A_3, S_{4}^{ex}) = (0.5, 1.9, 0.18, 0.15, 0.16, 0.1, 0.064)$ in the time interval $t\in[0, 3]$, assuming a ground truth set of parameters obtained from the experimental data provided in \cite{ruoff2003temperature}: $J_0 = 2.5$ mM/min, $k_1 = 100.0$ mM/min, $k_2 = 6.0$ mM/min, $k_3 = 16.0$ mM/min, $k_4 = 100.0$ mM/min, $k_5 = 1.28$ mM/min, $k_6 = 12.0$ mM/min, $k = 1.8$ mM/min, $\kappa = 13.0$ mM/min, $q = 4.0$, $K_I = 0.52$ mM, $N = 1.0$ nM, $A = 4.0$ mM and $\phi = 0.1$. The training dataset $\mathcal{D}$ is considered such that the simulated data is perturbed by $10\%$ white noise. Only $N_2$, $A_3$ and $S_{4}^{ex}$ are considered as observable variables for this example ($V=\{5,6,7\}$), but their corresponding observations are sampled at different time instances as follows:
\begin{equation}\label{eq:glyco_t}
    \begin{aligned}
        \bm{t}^{(5)}=(0.525,0.575,0.625,\ldots,2.925,2.975,3) \ , \\
        \bm{t}^{(6)}=(0,0.5,0.55,0.6,\ldots,2.9,2.95,3) \ , \\
        \bm{t}^{(7)}=(0,0.525,0.575,0.625,\ldots,2.925,2.975,3) \ .
    \end{aligned}
\end{equation}
Hence we have $51$ observations for variable $N_2$ with a time gap of $0.525$, a time step equal to $0.05$, and an unknown initial condition; $52$ observations for variable $A_3$ with a time gap of $0.5$ and a time step equal to $0.05$; and $52$ observations for variable $S_{4}^{ex}$ with a time gap of $0.525$ and a time step equal to $0.05$. For this example, the dynamics parameters vector is of size $15$ and given by $\bm{\theta_f}=(J_0,  k_1,  k_2, k_3, k_4, k_5, k_6, k, \kappa, q, K_I, \phi, N, A, N_{2,0})$, where $N_{2,0}$ is the inferred initial condition for the variable $N_2$. Given this noisy, sparse and irregular training data, our goal is to demonstrate the performance of the GP-NODE framework in identifying the unknown model parameters $\bm{\theta} = (\bm{\theta_f},\bm{\theta_g})$ (see equation (\ref{eq:glob_par})) by inferring their posterior distribution. The prior distribution of $\bm{\theta_g}$ is discussed in section \ref{sec:normalization}, and the prior distributions of $\bm{\theta_f}$ are taken as the uniform distributions defined over the intervals detailed in table \ref{tab:Glycolysis_priors}.

\begin{table}
\centering
\begin{tabular}{|c|c|c|c|c|c|}
\hline
 & $J_0$ & $k_1$ & $k_2$ & $k_3$ & $k_4$ \\ 
\hline
Interval  & [1,10] & [80,120] & [1,10] & [2,20] & [80,120] \\
\hline 
& $k_5$ & $k_6$ & $k$ & $\kappa$ & $q$ \\
\hline
Interval  & [0.1,2] & [2,20] & [0.1,2] & [2,20] & [1,10] \\
\hline
& $K_I$ & $\phi$ & $N$ & $A$ & $N_{2,0}$  \\
\hline
Interval  & [0.1,2] & [0.05,1] & [0.1,2] & [1,10] & [0,1] \\
\hline
\end{tabular}
\caption{{\em Yeast Glycolysis dynamics:} Interval of uniform distributions considered as priors for parameters $\bm{\theta_f}$.}
\label{tab:Glycolysis_priors}
\end{table}

\begin{figure}
\centering
\includegraphics[width=\textwidth]{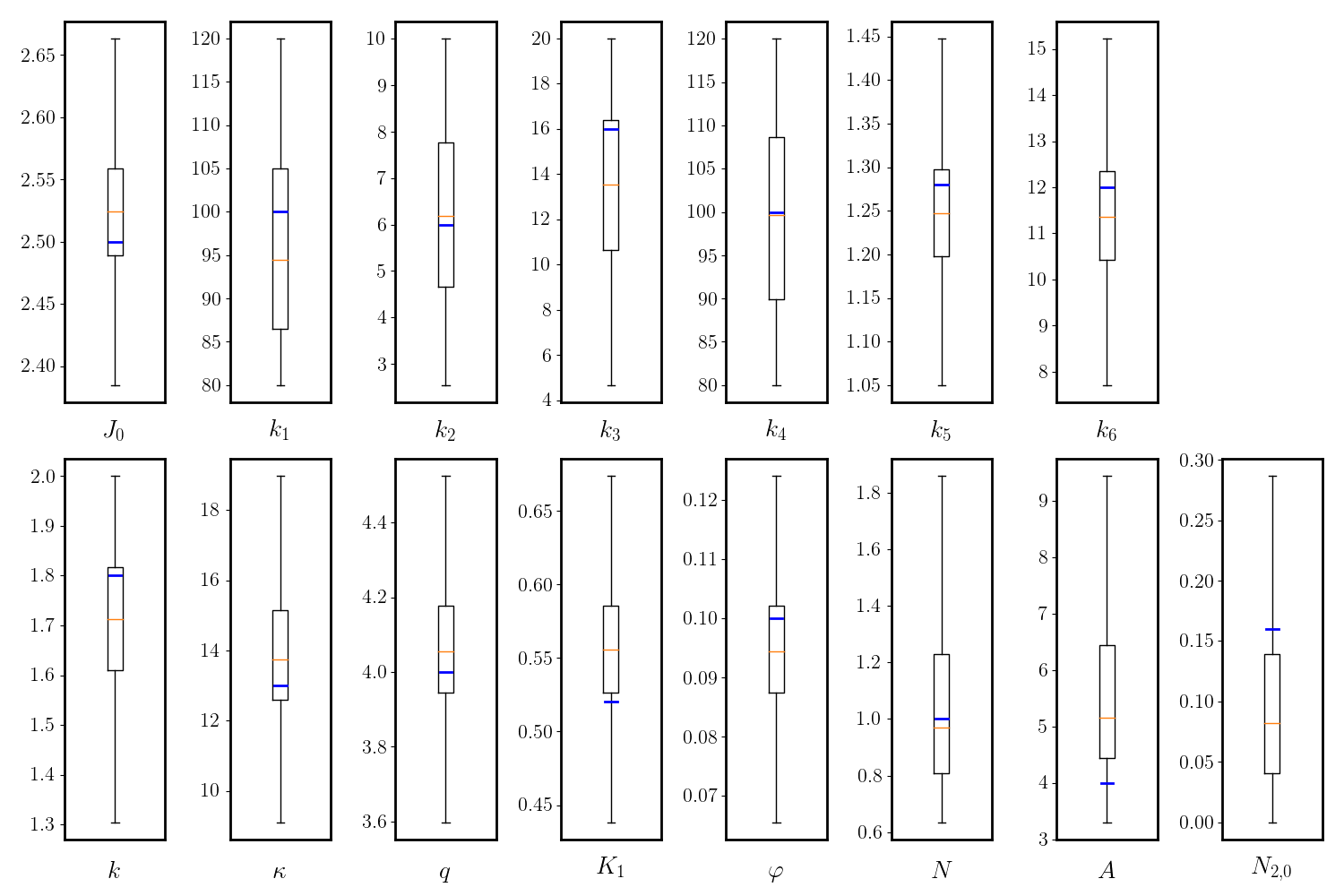}
\caption{{\em Yeast Glycolysis dynamics:} Uncertainty estimation of the inferred model parameters obtained using the proposed Bayesian differential programming method. Estimates for the minimum, maximum, median, first quantile and third quantile are provided, while the true parameter values are highlighted in blue.}
\label{fig:12k_glyco_LV_0123_t_gap_diff_t_x0_test_no_crop}
\end{figure}

Uncertainty estimates for the inferred parameters can be deducted from the computed posterior distribution $p(\bm{\theta_f}|\mathcal{D})$, as presented in the box plots of figure \ref{fig:12k_glyco_LV_0123_t_gap_diff_t_x0_test_no_crop} where the minimum, maximum, median, first quantile and third quantile obtained from the HMC simulations for each parameter are presented. All inferred parameters closely agree with the ground truth values used to generate the training data as reported in \cite{ruoff2003temperature}. Notice that that all true values fall between the predicted quantiles, while for a few parameters (only $K_1$, $A$ and $N_{2,0}$ out of the total $15$ parameters considered), the true values are not perfectly captured by the confidence intervals. This behavior can be explained by the lower sensitivity of the dynamical trajectories with respect to those parameters, and also by the relatively small size of the sparse training data considered since only $3$ variables of the 7-dimensional dynamical system are observed with one unknown initial condition. 

\begin{figure}
\centering
\includegraphics[width=\textwidth]{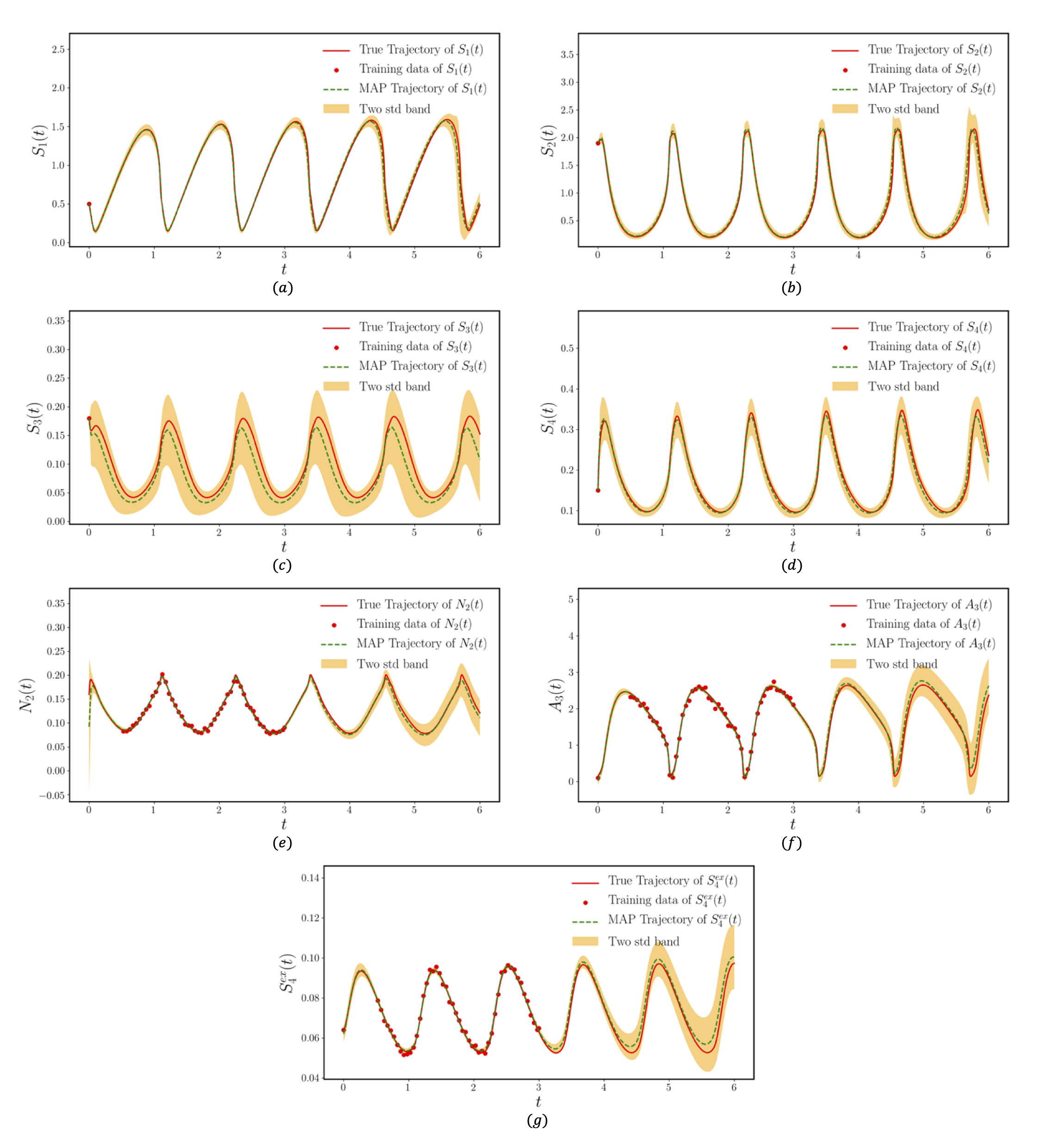}
\caption{{\em Learning Yeast Glycolysis dynamics:} (a) Learned dynamics versus the true dynamics and the training data of $S_1$. (b) Learned dynamics versus the true dynamics and the training data of $S_2$. (c) Learned dynamics versus the true dynamics and the training data of $S_3$. (d) Learned dynamics versus the true dynamics and the training data of $S_4$. (e) Learned dynamics versus the true dynamics and the training data of $N_2$. (f) Learned dynamics versus the true dynamics and the training data of $A_3$. (g) Learned dynamics versus the true dynamics and the training data of $S_4^{ex}$.}
\label{fig:12k_glyco_LV_0123_t_gap_diff_t_x0}
\end{figure}

The inferred model also allows us to produce forecasts of future states with quantified uncertainty via the predictive posterior distribution of equation (\ref{equ:predictive_distribution}). These extrapolated states are shown in figure \ref{fig:12k_glyco_LV_0123_t_gap_diff_t_x0}. It is evident that the uncertainty estimation obtained via sampling from the joint posterior distribution over all model parameters $p(\bm{\theta_f},\bm{\theta_g}|\mathcal{D})$ is able to well capture the reference trajectory of the high dimensional dynamical system. As expected, the uncertainty is larger for latent variables, such as $S_3$, and for unseen time instances. Nonetheless, even for these cases, the uncertainty estimates always capture the exact trajectory and the MAP trajectory follows the true one with sufficient accuracy.

\begin{figure}
\centering
\includegraphics[width=\textwidth]{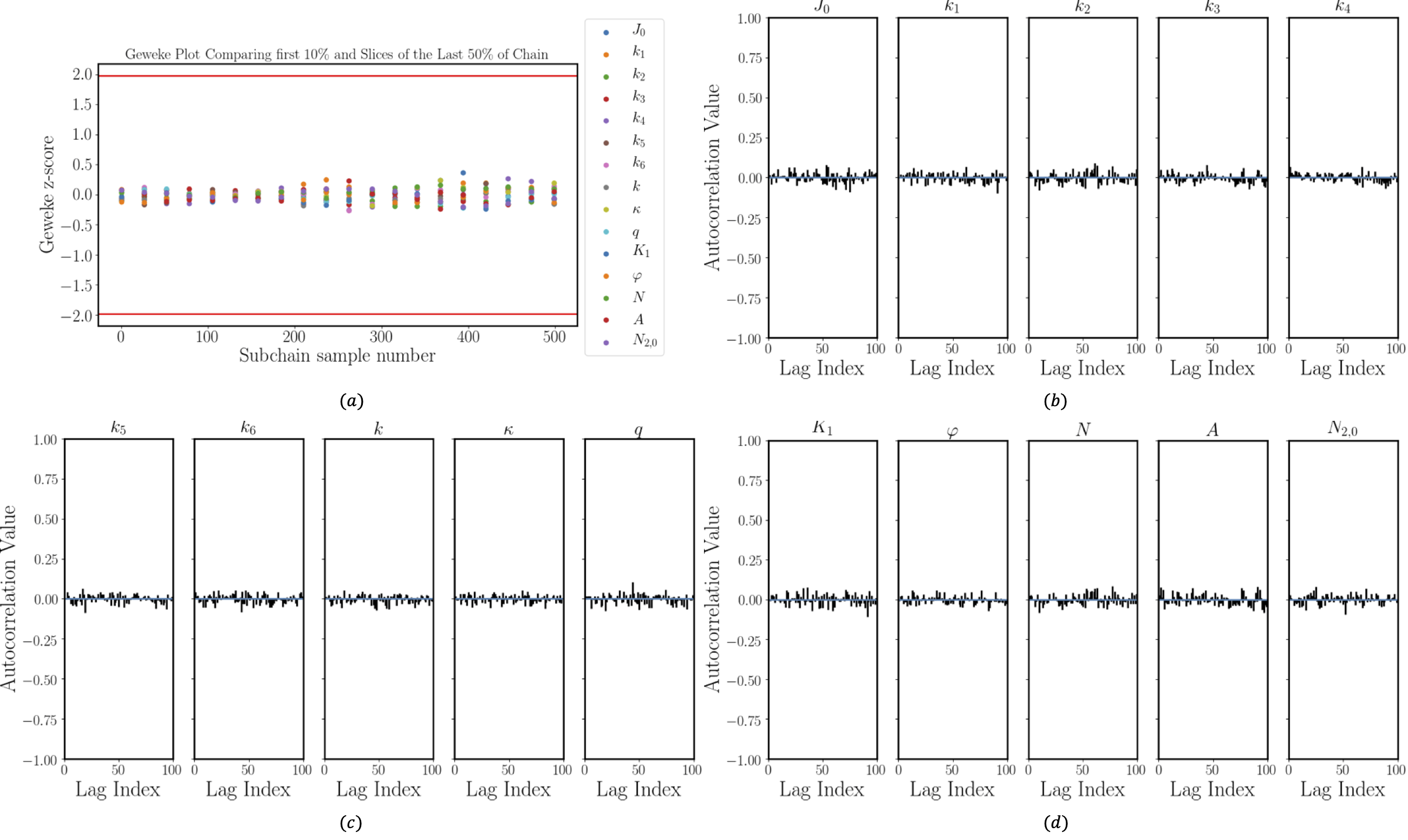}
\caption{\textcolor{black}{{\em HMC convergence diagnostics for the Yeast-Glycolysis system parameter $\bm{\theta_f}$:} (a) The Geweke diagnostic test based on the chain of $1000$ samples: we compare the mean of the first  10\% of the chain of $1000$ samples to the mean of $20$ sub-samples from the last 50\% of the chain. (b), (c) and (d) Auto-correlation as function of the lag.}}
\label{fig:glyco_MCMC_diagnostics}
\end{figure}

To verify the statistical convergence of the HMC chains, we have employed the Geweke diagnostic and Gelman Rubin tests \cite{geweke1993bayesian,gelman2013bayesian}. The Gelman Rubin tests have provided the following values for the parameter $\bm{\theta_f}$ ($J_0: 1.034,  k_1: 1.018,  k_2: 1.035, k_3: 1.165, k_4: 1.029, k_5: 1.390, k_6: 1.607, k: 1.066, \kappa: 1.247, q: 1.187, K_I: 1.006, \phi: 1.274, N: 1.005, A: 1.016, N_{2,0}: 1.407$), which are all close to the optimal value of $1$, as expected. In order to perform the Geweke diagnostic, we have ``thinned" the $8,000$ HMC draws by randomly sampling one data point every eight consecutive steps of the chain. In order to implement the Geweke diagnostic, we considered the first 10\% of the chain of the $1,000$ samples and compared their mean with the mean of $20$ sub-samples from the last 50\% of the chain. As shown in plot (a) of figure \ref{fig:glyco_MCMC_diagnostics}, the Geweke z-scores for the four non-zero parameters are well between $-2$ and $2$. The results of auto-correlation are given in plots (b), (c) and (d) of figure \ref{fig:glyco_MCMC_diagnostics}. The Geweke diagnostic, the Gelman Rubin tests results and the auto-correlation values demonstrate a ``healthy" mixing behavior of the NUTS sampler.

\begin{figure}
\centering
\includegraphics[width=\textwidth]{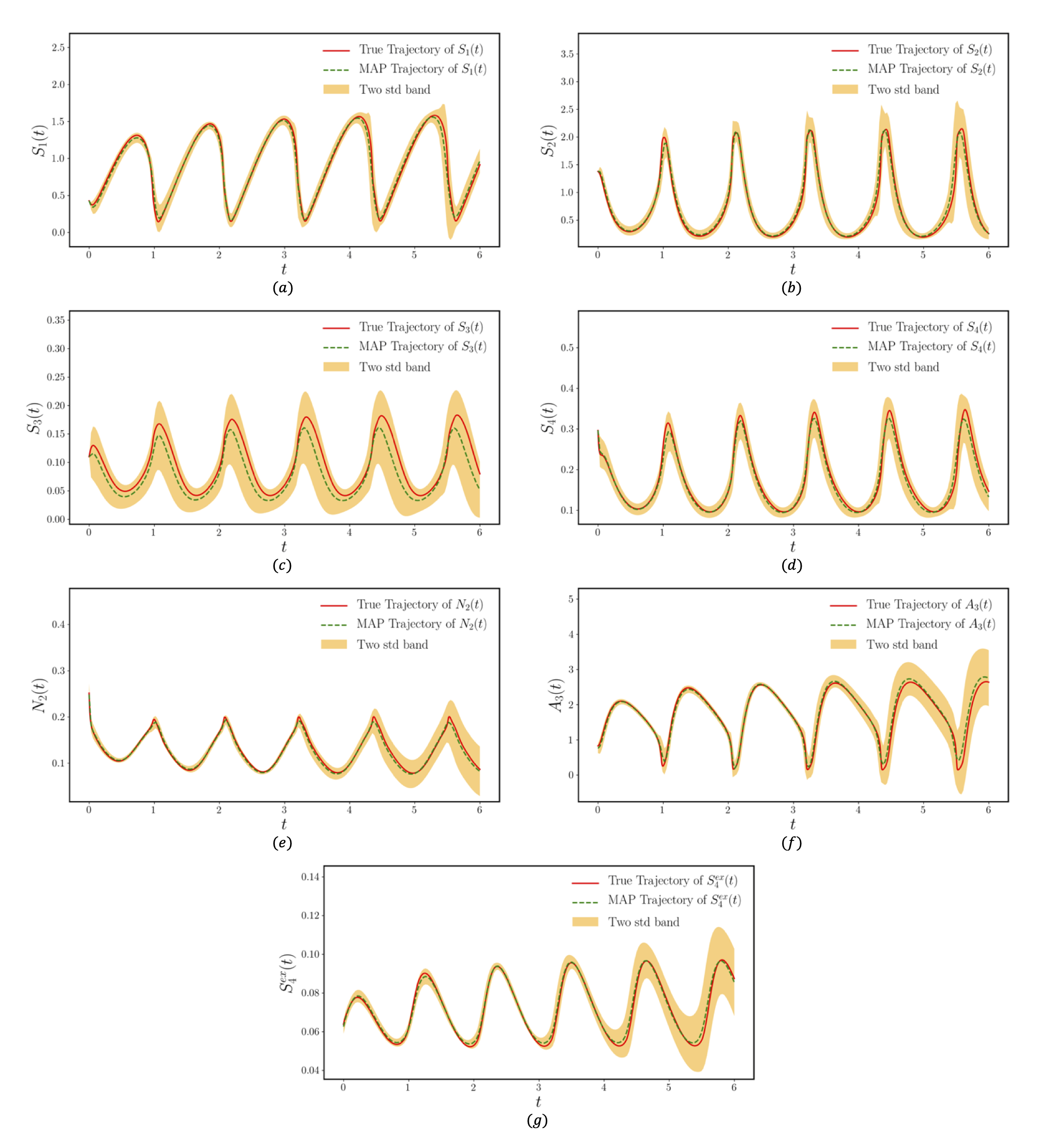}
\caption{{\em Learning Yeast Glycolysis dynamics:} Future forecasts with quantified uncertainty from unseen initial conditions that were not used during model training.}
\label{fig:random_x0_12k_glyco_LV_0123_t_gap_diff_t_x0}
\end{figure}

Finally, to illustrate the generalization capability of the inferred model with respect to different initial conditions than those used during training, we have assessed the quality of the predicted states starting from a random initial condition that is uniformly sampled within ($S_1\in[0.15,1.60], S_2\in[0.19,2.10], S_3\in[0.04,0.20], S_4\in[0.10,0.35], N_2\in[0.08,0.30], A_3\in[0.14,2.67], S_{4}^{ex}\in[0.05,0.10]$). Figure \ref{fig:random_x0_12k_glyco_LV_0123_t_gap_diff_t_x0} shows the prediction obtained with the randomly picked initial condition $[0.428, 1.378, 0.110, 0.296, 0.252, 0.830, 0.064]$. 
% Note that the model was trained using the trajectory obtained with the previously specified initial condition:  $[S_1, S_2, S_3, S_4, N_2, A_3, S_{4}^{ex}] = [0.5, 1.9, 0.18, 0.15, 0.16, 0.1, 0.064]$). 
The close agreement with the reference solution indicates that the inferred model is well capable of generalizing both in terms of handling different initial conditions, as well as extrapolating to reliable future forecasts with quantified uncertainty.

\subsection{Dictionary learning of human motion capture data}\label{sec:hum_mot}
In this final example, our goal is to investigate the performance of the GP-NODE algorithm in a realistic problem using experimental data. To this end, we consider a benchmark data-set of human motion capture data from the Carnegie Mellon University motion capture (CMU mocap) database. The data-set contains $50$-dimensional pose measurements of human walking motion that here we denote as $\bm{y}(t_i)$. Each pose dimension records a measurement in different parts of the body during movement \cite{Wang2008}. Following previous studies on human motion capture \cite{Wang2006,Wang2008,Damianou2011,Heinonen2018}, we perform Principal Component Analysis (PCA) in order to project the original data-set to a latent space spanning the 3 dominant eigen-directions. Hence, the state space considered $\bm{x}(t)$ is three dimensional, and the original $50$-dimensional state $\bm{y}(t)$ can be recovered through the corresponding PCA projection.

The performance of the proposed method is evaluated by removing around $20\%$ of the frames from the middle of the trajectories. The latter will be used to assess the predictive accuracy of the method. The goal is to learn a model with the remaining data, and to forecast the missing values. We measure the root mean square error (RMSE) over the original feature space of the 50-dimensional state $\bm{y}(t)$ for which the experimental measurements are provided. The original dimensions are reconstructed from the latent space trajectories of $\bm{x}(t)$. We compare our method to the recently proposed Nonparametric ODE Model method (npODE) \cite{Heinonen2018}, which was shown to outperform the Gaussian Process Dynamical Model approach (GPDM) \cite{Wang2006} and the Variational Gaussian Process Dynamical System method (VGPDS) \cite{Damianou2011}. The npODE results are obtained using the original implementation provided by the authors at \url{https://github.com/cagatayyildiz/npode}.

First, we apply the proposed Bayesian approach to recover the dynamical system of human motion data using a polynomial dictionary parameterization taking the form:

\begin{equation} \label{eq:dict_A}
\begin{bmatrix}
    \frac{dx_1}{dt}\\
    \frac{dx_2}{dt}
\end{bmatrix}
= A\varphi(\bm{x}) = 
\begin{bmatrix}
  a_{11} & a_{12} & a_{13} & a_{14} & a_{15} & a_{16} & a_{17} \\
  a_{21} & a_{22} & a_{23} & a_{24} & a_{25} & a_{26} & a_{27} \\
  a_{31} & a_{32} & a_{33} & a_{34} & a_{35} & a_{36} & a_{37}
\end{bmatrix}
\begin{bmatrix}
   1 \\ x_1 \\ x_2 \\ x_3 \\ x_1 x_2 \\ x_1 x_3 \\ x_2 x_3
\end{bmatrix}: \ {\rm case A} \ ,
\end{equation}
where the $a_{ij}$'s are unknown scalar coefficients that will be estimated.

We consider the trajectory ``07\_02.amc" from the CMU mocap database, such that we have a total of $82$ pose measurements $\bm{y}(t_i)$. The frames from the $35$-th to the $49$-th measurements are removed, such that the model is trained with $67$ observations. For this example, the latent dynamics parameter vector is simply given by $\bm{\theta_f}=\bm{\theta_A}$, where $\bm{\theta_A}$ is the vector containing the coefficients of $A$. Given the truncated training data, the GP-NODE framework is used to identify the unknown model parameters $\bm{\theta} = (\bm{\theta_f},\bm{\theta_g})$ (see equation (\ref{eq:glob_par})) by inferring their posterior distributions. The Finnish Horseshoe distribution is used as prior for the parameters $\bm{\theta_A}$ as discussed in section \ref{sec:learning_dynamics_dict}, while the prior distribution of $\bm{\theta_g}$ is detailed in section \ref{sec:normalization}.

\begin{figure}
\centering
\includegraphics[width=\textwidth]{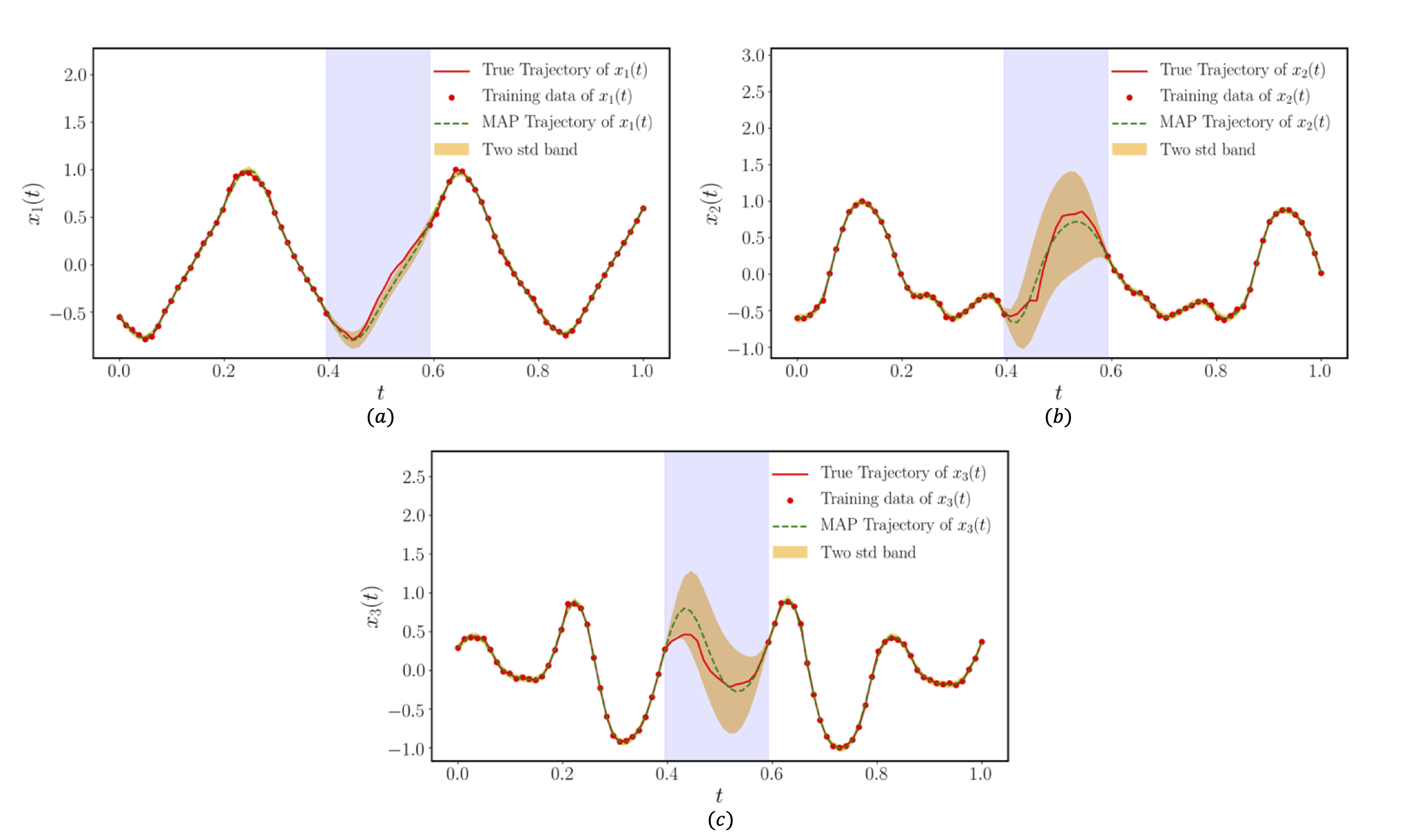}
\caption{{\em Parameter inference of a human motion dynamics system, case A:} (a) Learned dynamics versus the true dynamics and the training data of $x_1(t)$. (b) Learned dynamics versus the true dynamics and the training data of $x_2(t)$. (c) Learned dynamics versus the true dynamics and the training data of $x_3(t)$. The shaded region indicates the time range of unseen data during model training.}
\label{fig:time_matlab_A_m0_20_check_robot_uncorr_finnish_hs_12k_GP}
\end{figure}

Our numerical results are summarized in figure \ref{fig:time_matlab_A_m0_20_check_robot_uncorr_finnish_hs_12k_GP} showing the training data and the estimated trajectories based on the inferred parameters. It is evident that the proposed methodology is able to provide an estimator with a predicted trajectory that closely matches the exact system's dynamics even for the unseen data. Indeed, for the time interval corresponding to the missing data, the MAP trajectories for the different variables $\bm{x}_i$ follow the true ones. Moreover the true trajectories always reside within the two standard deviations band, showing that the method is able to determine a posterior distribution over plausible models that captures the epistemic uncertainty of the assumed parametrization and the uncertainty induced by training on a truncated and noisy training data. Such uncertainty is propagated through the system dynamics to characterize variability in the unseen system states, since the highest uncertainty is obtained for the time interval corresponding to missing data.

\begin{figure}
\centering
\includegraphics[width=\textwidth]{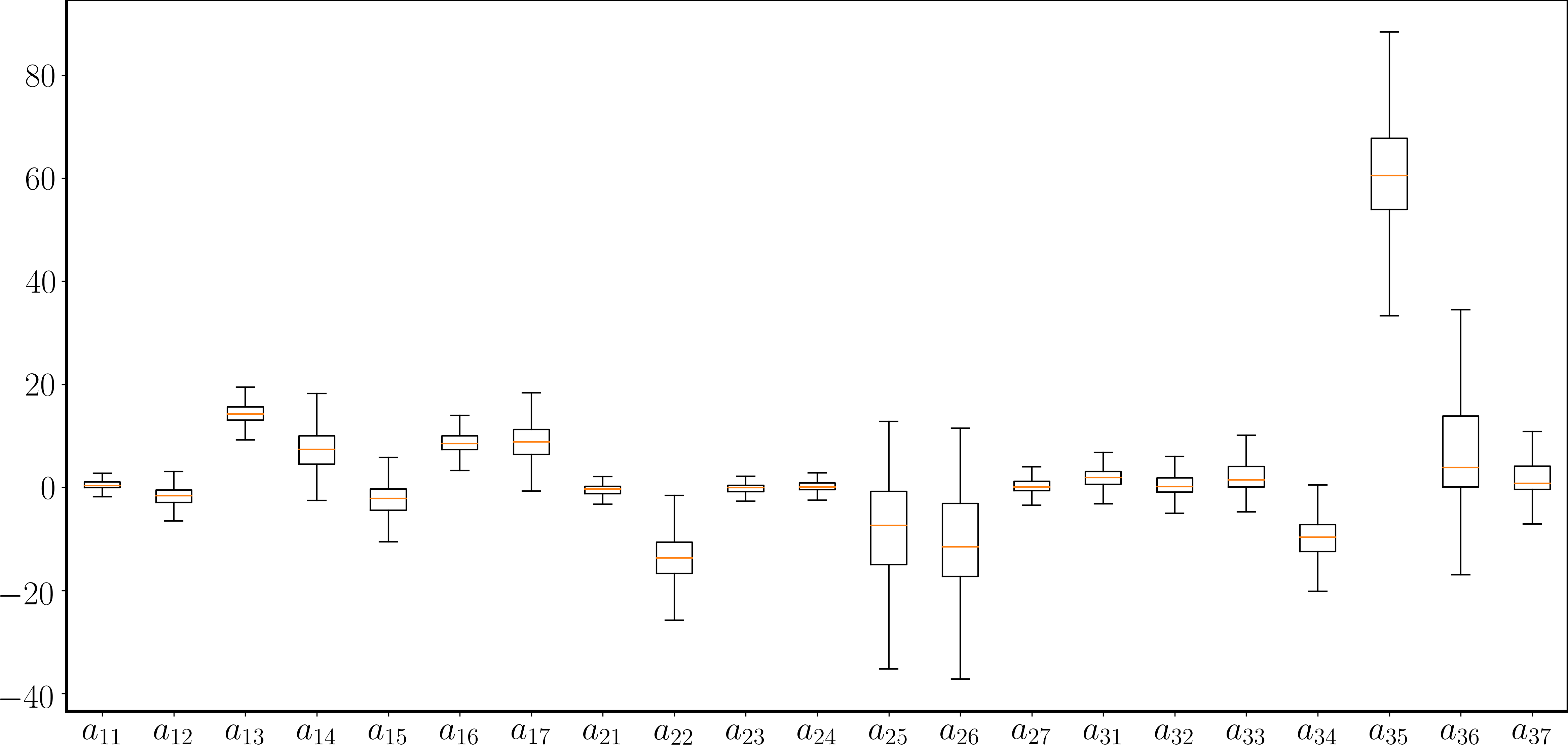}
\caption{{\em Human motion dynamics system, case A:} Uncertainty estimation of the inferred model parameters obtained using the GP-NODE method. Estimates for the minimum, maximum, median, first quantile and third quantile are provided.}
\label{fig:time_matlab_A_m0_20_check_robot_uncorr_finnish_hs_12k_no_crop}
\end{figure}

\begin{figure}
\centering
\includegraphics[width=\textwidth]{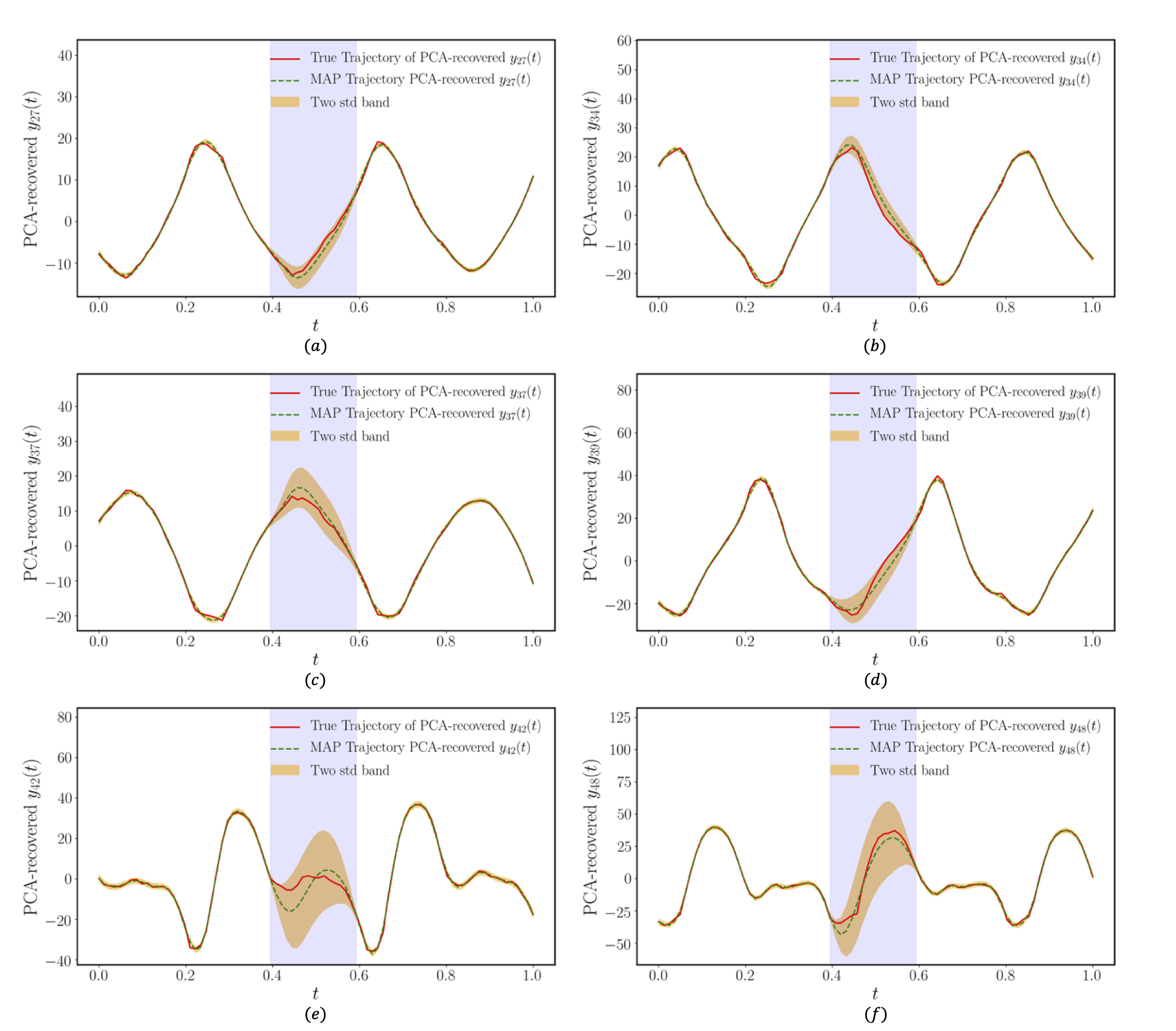}
\caption{{\em Parameter inference of a human motion dynamics system, case A:} (a) Learned dynamics versus the true dynamics of PCA-recovered $y_{27}(t)$. (b) Learned dynamics versus the true dynamics of PCA-recovered $y_{34}(t)$. (c) Learned dynamics versus the true dynamics of PCA-recovered $y_{37}(t)$. (d) Learned dynamics versus the true dynamics of PCA-recovered $y_{39}(t)$. (e) Learned dynamics versus the true dynamics of PCA-recovered $y_{42}(t)$. (f) Learned dynamics versus the true dynamics of PCA-recovered $y_{48}(t)$. The shaded region indicates the time range of unseen data during model training.}
\label{fig:time_matlab_A_m0_20_check_robot_uncorr_finnish_hs_12k_Y_GP}
\end{figure}

Uncertainty estimates for the inferred parameters can also be deducted from the computed posterior distribution $p(\bm{\theta_f}|\mathcal{D})$, as presented in the box plots of figure \ref{fig:time_matlab_A_m0_20_check_robot_uncorr_finnish_hs_12k_no_crop} where the minimum, maximum, median, first quantile and third quantile obtained from the MCMC simulations for each parameter are presented. We also assess the performance of the method by verifying the accuracy of the predicted $50$-dimensional state $\bm{y}(t)$ that are recovered via PCA reconstruction. These numerical results are given in figure \ref{fig:time_matlab_A_m0_20_check_robot_uncorr_finnish_hs_12k_Y_GP} showing the obtained prediction for the PCA-recovered trajectories of $6$ of the most dominant variables of the $50$-dimensional state, compared to the corresponding true trajectories. As observed for the state $\bm{x}$, the MAP trajectories for the PCA-recovered variables follow the true ones, and the uncertainty is well quantified for the time interval corresponding to missing data.

\begin{figure}
\centering
\includegraphics[width=\textwidth]{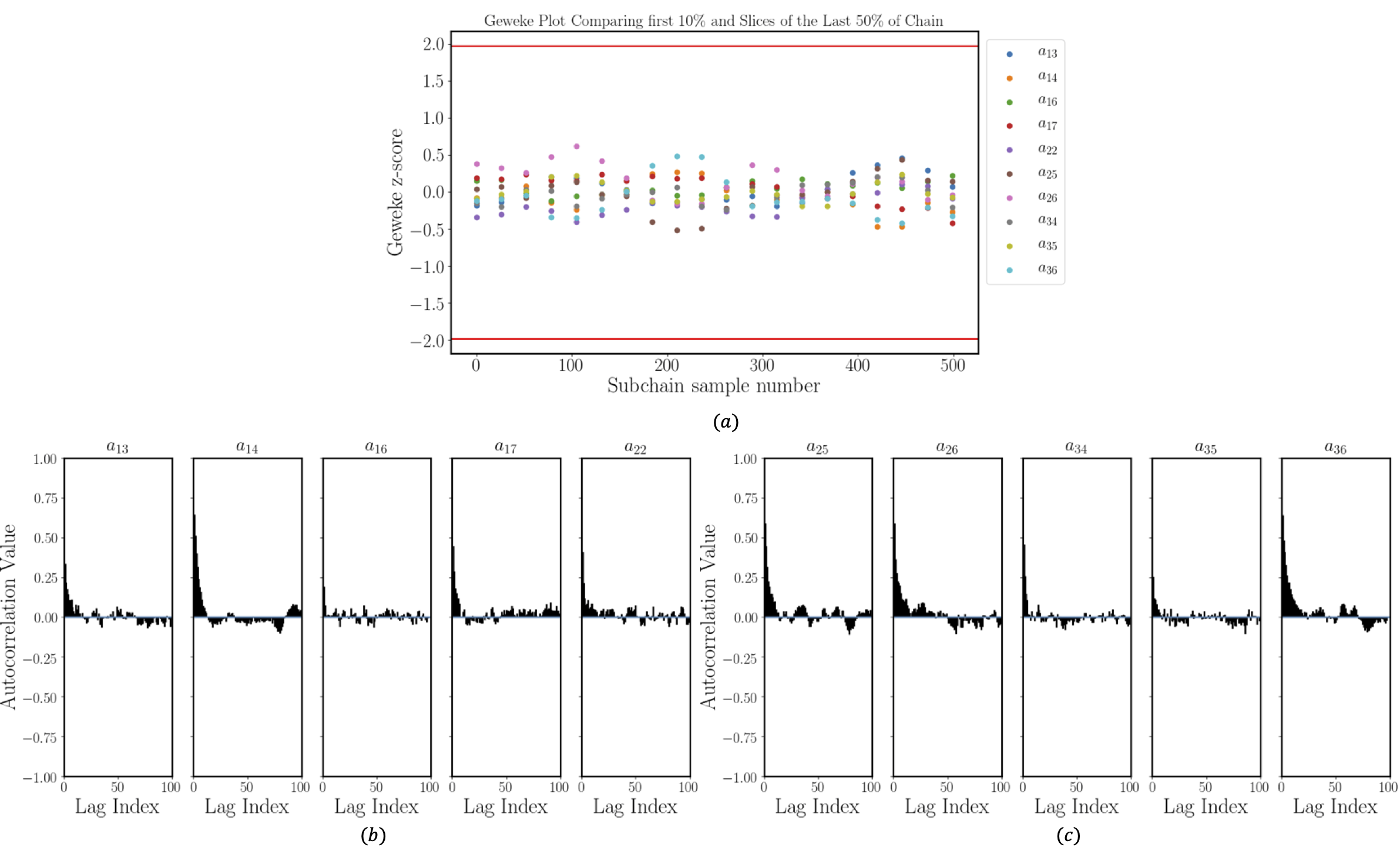}
\caption{\textcolor{black}{{\em HMC convergence diagnostics for most significant parameters of human motion system:} (a) The Geweke diagnostic test based on the chain of $1000$ samples: we compare the mean of the first  10\% of the chain of $1000$ samples to the mean of $20$ sub-samples from the last 50\% of the chain. (b) and (c) Auto-correlation as function of the lag.}}
\label{fig:robot_MCMC_diagnostic}
\end{figure}

To verify the statistical convergence of the HMC chains, we have again employed the Geweke diagnostic and Gelman Rubin tests \cite{geweke1993bayesian,gelman2013bayesian}. The Gelman Rubin tests have provided the following values for the most significant parameters of $\bm{\theta_f}$ ($a_{13}: 1.001, a_{14}: 1.004, a_{16}: 1.000, a_{17}: 1.000, a_{22}: 1.003, a_{25}: 1.001, a_{26}: 1.006, a_{34}: 1.001, a_{35}: 1.001, a_{36}: 1.002$), which are all close to the optimal value of $1$, as expected. In order to perform the Geweke diagnostic, we have ``thinned" the $8,000$ HMC draws by randomly sampling one data point every eight consecutive steps of the chain. In order to implement the Geweke diagnostic, we considered the first 10\% of the chain of the $1000$ samples and compared their mean with the mean of $20$ sub-samples from the last 50\% of the chain. As shown in plot (a) of figure \ref{fig:robot_MCMC_diagnostic}, the Geweke z-scores for the four non-zero parameters are well between $-2$ and $2$. The results of auto-correlation are given in plots (b) and (c) of figure \ref{fig:robot_MCMC_diagnostic}. The Geweke diagnostic, the Gelman Rubin tests results and the auto-correlation values demonstrate a ``healthy" mixing behavior of the NUTS sampler.

In order to demonstrate the robustness of the proposed method with respect to the size of the dictionary, we augment the previous dictionary (see equation \ref{eq:dict_A}) to include a complete second order polynomial approximation for the latent dynamics as follows:
\begin{equation} \label{eq:dict_B}
\begin{bmatrix}
    \frac{dx_1}{dt}\\
    \frac{dx_2}{dt}
\end{bmatrix}
= A\varphi(\bm{x}) = 
\begin{bmatrix}
  a_{11} & a_{12} & a_{13} & a_{14} & a_{15} & a_{16} & a_{17} & a_{18} & a_{19} & a_{110} \\
  a_{21} & a_{22} & a_{23} & a_{24} & a_{25} & a_{26} & a_{27} & a_{28} & a_{29} & a_{210} \\
  a_{31} & a_{32} & a_{33} & a_{34} & a_{35} & a_{36} & a_{37} & a_{38} & a_{39} & a_{310}
\end{bmatrix}
\begin{bmatrix}
   1 \\ x_1 \\ x_2 \\ x_3 \\ x_1 x_2 \\ x_1 x_3 \\ x_2 x_3 \\ x_1^2 \\ x_2^2 \\ x_3^2
\end{bmatrix}: \ {\rm case B} \ ,
\end{equation}
where the $a_{ij}$'s are unknown scalar coefficients that will be estimated.

\begin{figure}
\centering
\includegraphics[width=\textwidth]{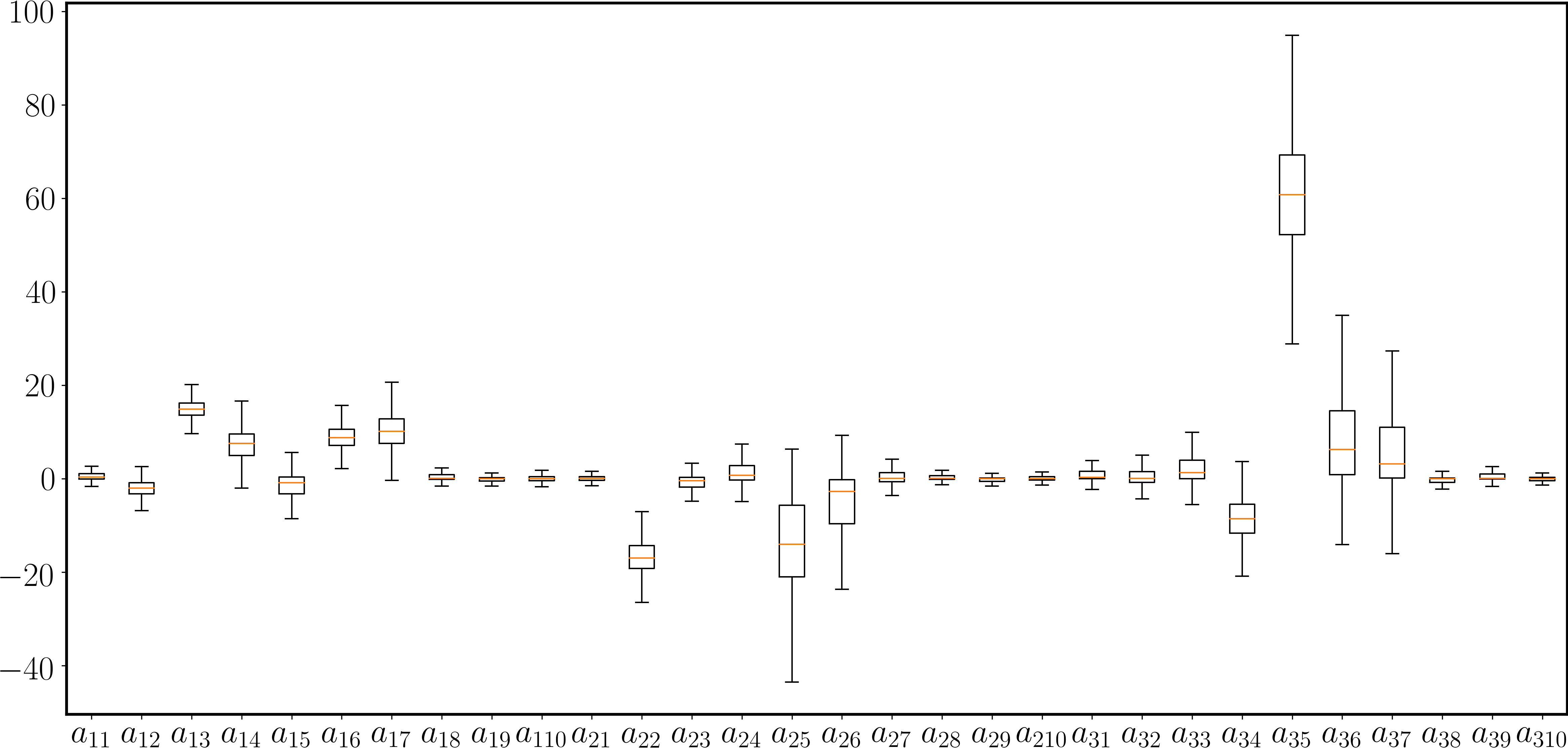}
\caption{{\em Human motion dynamics system, case B:} Uncertainty estimation of the inferred model parameters obtained using the GP-NODE method. Estimates for the minimum, maximum, median, first quantile and third quantile are provided.}
\label{fig:time_matlab_B_m0_29_check_robot_uncorr_finnish_hs_12k_no_crop}
\end{figure}

We consider the same training data-set and prior distributions as detailed above for case A. Uncertainty estimates for the inferred parameters are presented in the box plots of figure \ref{fig:time_matlab_B_m0_29_check_robot_uncorr_finnish_hs_12k_no_crop} where the minimum, maximum, median, first quantile and third quantile obtained from the HMC simulations for each parameter are presented. Interestingly, based on the estimates detailed in figure \ref{fig:time_matlab_A_m0_20_check_robot_uncorr_finnish_hs_12k_no_crop}, it is clear that the method is robust with respect to the chosen dictionary and does provide close estimates for the parameters $a_{ij}$ that correspond to the same terms in the two dictionaries, while it assigns zero values to the ones corresponding to the additional terms added to the dictionary of case B.

\begin{figure}
\centering
\includegraphics[width=\textwidth]{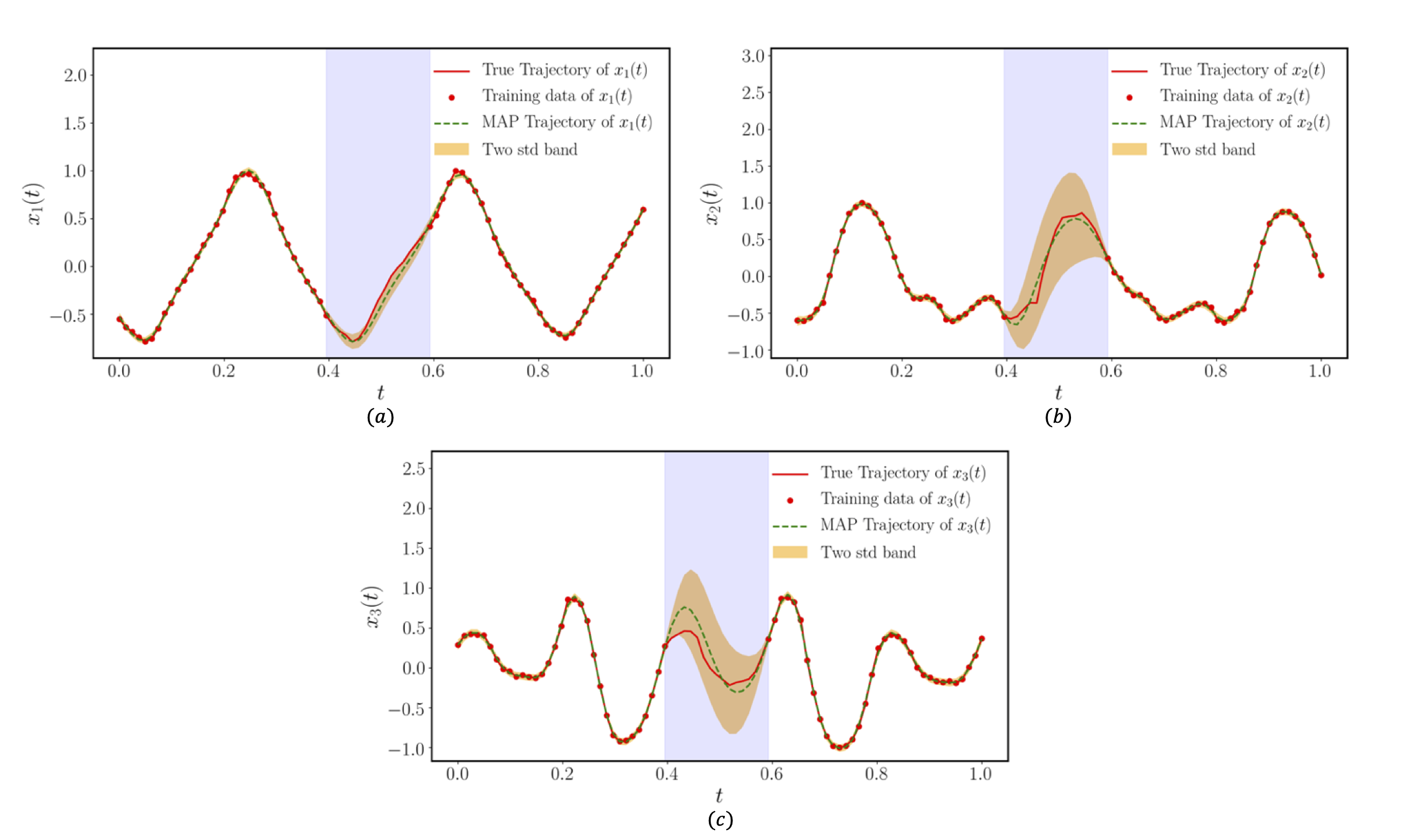}
\caption{{\em Parameter inference of a human motion dynamics system, case B:} (a) Learned dynamics versus the true dynamics and the training data of $x_1(t)$. (b) Learned dynamics versus the true dynamics and the training data of $x_2(t)$. (c) Learned dynamics versus the true dynamics and the training data of $x_3(t)$. The shaded region indicates the time range of unseen data during model training.}
\label{fig:time_matlab_B_m0_29_check_robot_uncorr_finnish_hs_12k_GP}
\end{figure}

\begin{figure}
\centering
\includegraphics[width=\textwidth]{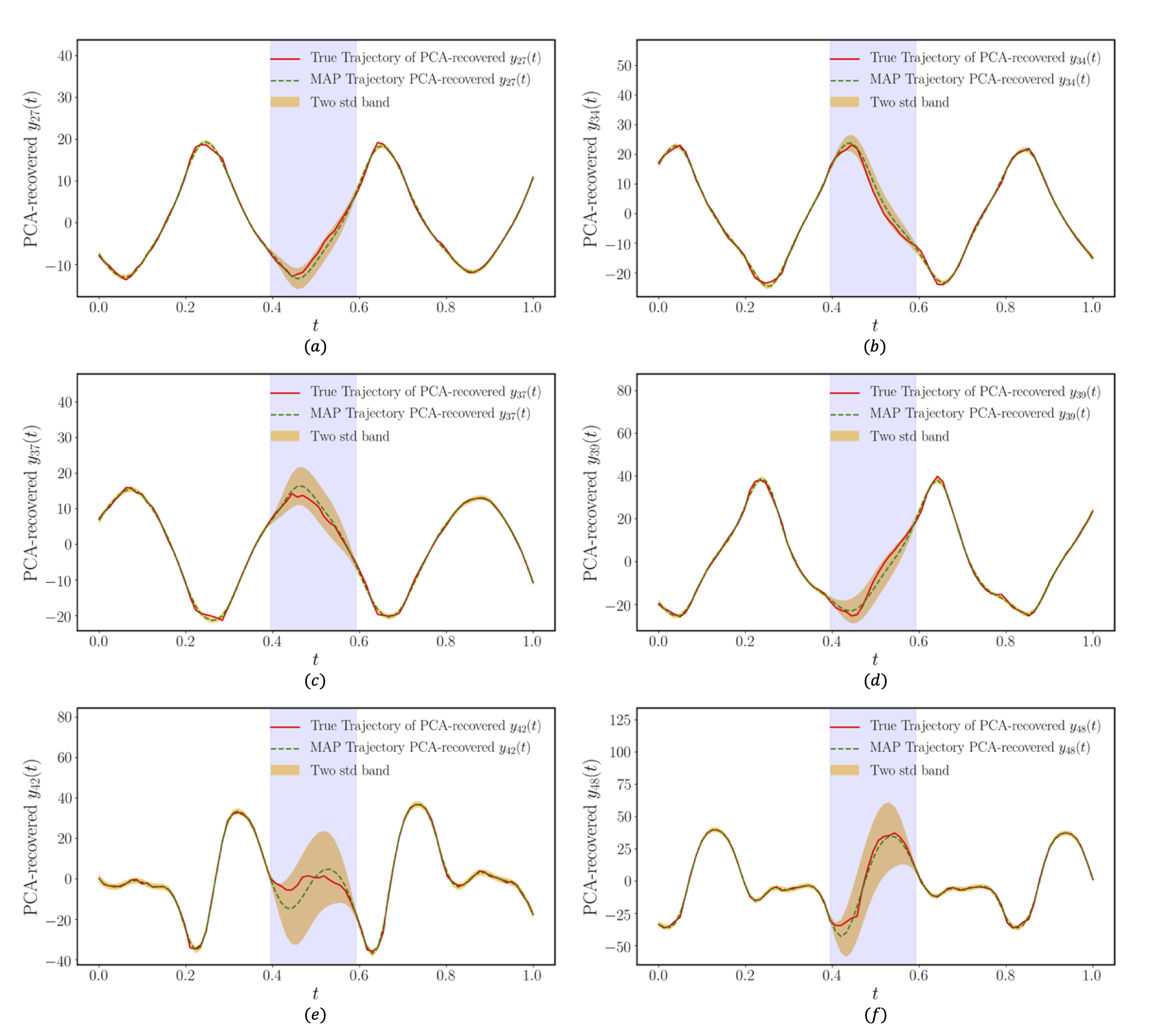}
\caption{{\em Parameter inference of a human motion dynamics system, case B:} (a) Learned dynamics versus the true dynamics of PCA-recovered $y_{27}(t)$. (b) Learned dynamics versus the true dynamics of PCA-recovered $y_{34}(t)$. (c) Learned dynamics versus the true dynamics of PCA-recovered $y_{37}(t)$. (d) Learned dynamics versus the true dynamics of PCA-recovered $y_{39}(t)$. (e) Learned dynamics versus the true dynamics of PCA-recovered $y_{42}(t)$. (f) Learned dynamics versus the true dynamics of PCA-recovered $y_{48}(t)$. The shaded region indicates the time range of unseen data during model training.}
\label{fig:time_matlab_B_m0_29_check_robot_uncorr_finnish_hs_12k_Y_GP}
\end{figure}

As observed for case A, the MAP trajectories obtained with the case B dictionary for the state $\bm{x}$ and for the PCA-recovered variables follow the true ones, and the uncertainty is also well quantified for the time interval corresponding to missing data as shown in figures \ref{fig:time_matlab_B_m0_29_check_robot_uncorr_finnish_hs_12k_GP} and \ref{fig:time_matlab_B_m0_29_check_robot_uncorr_finnish_hs_12k_Y_GP}. Table \ref{tab:human_motion_RMSE} provides the RMSE between the original $50$-dimensional state $\bm{y}(t)$ for which the experimental measurements are provided, and the PCA-recovered $50$-dimensional state obtained with the GP-NODE approach for cases A and B. The errors obtained using npODE \cite{Heinonen2018} are also detailed in table \ref{tab:human_motion_RMSE}, showing that the GP-NODE outperforms npODE not only in fitting the observed data, but also on forecasting the missing region.

\begin{table}
\centering
\begin{tabular}{|c|c|c|}
\hline
Model & Fitting observed data & Forecasting missing data \\ 
\hline
npODE & $6.89 $ & $9.79 $ \\
\hline 
GP-NODE, case A & $\bm{3.14\pm 0.006}$ & $\bm{4.67\pm 0.65}$ \\
\hline
GP-NODE, case B & $\bm{3.14\pm 0.005}$ & $\bm{4.49\pm 0.60}$ \\
\hline 
\end{tabular}
\caption{{\em Human motion dynamics system:} RMSE in fitting observed data and forecasting missing data.}
\label{tab:human_motion_RMSE}
\end{table}

\section{Conclusions}\label{sec:discussion}

We put forth a novel machine learning framework (GP-NODE) for robust systems identification under uncertainty and imperfect time-series data. The GP-NODE framework leverages state-of-the-art differential programming techniques in combination with sparsity promoting priors, Gaussian processes and gradient-enhanced sampling schemes for scalable Bayesian inference of high-dimensional posterior distributions, which allows an inference of interpretable and parsimonious representations of complex dynamical systems with end-to-end uncertainty quantification. The GP-NODE workflow can naturally accommodate sparse and noisy time-series data; latent variables; unknown initial conditions; irregular time sampling for each observable variable; and observations at different time instances for the observable variables.  The effectiveness of the GP-NODE technique has been systematically investigated and compared to state-of-the-art approaches across different problems including synthetic numerical examples and model discovery from experimental data for human motion. These applications show how the GP-NODE probabilistic formulation is robust against erroneous data, incomplete model parametrization, and produces reliable future forecasts with uncertainty quantification.

% high-dimensional Bayesian inference, the proposed Bayesian \textcolor{red}{differential programming framework} accelerates the model parameters discovery with reliable uncertainties. Two different contexts were considered: (1) a data driven discovery of dynamical system context with a dictionary learning task, which is illustrated by a damped oscillator system, \textcolor{red}{a damped pendulum system} and a Lorenz system, and (2) a parameter identification context for dynamical systems with an available physics knowledge which is demonstrated by the prey-predator dynamical system and a realistic high dimensional biological system. For both contexts, the training data considered is irregular and noisy. The proposed method has the capability to tackle uncertainty quantification tasks for large set of unknown parameters in high dimensional dynamical systems with irregular and noisy observations. 

Although the GP-NODE framework offers great flexibility to infer a distribution over plausible parsimonious representations of a dynamical system, a number of technical improvements can be further pursued. The first relates to devising more effective initialization procedures for Hamiltonian Monte Carlo sampling since the initial estimations of the system parameters may cause the system to become stiff, especially for relatively sparse observations, which can lead to numerical instabilities during model training. Such issue can be tackled by adopting stiffly stable ODE solvers \cite{rackauckas2020universal, gholami2019anode}. A second direction of generalization could be in applying the GP-NODE method to parameter identification of partial differential equations (PDEs) with the appropriate improvements, since discretization of PDEs generally translates into high dimensional dynamical systems. Optimization of the discretization scheme can be investigated within such a context \cite{rackauckas2020universal}. Finally, one could try to generalize the GP-NODE approach to stochastic problems where the dynamics itself are driven by a stochastic process. The approaches developed in \cite{li2020scalable} could be useful for instance.

%Although a series of promising empirical results were presented, we would like to also point out that a rigorous theoretical justification on the convergence of the proposed methods is still missing. To our understanding there are two fundamental concepts that need to be investigated: (i) under which conditions a dynamical system is identifiable from raw trajectory data, and (ii) the consistency of Bayesian inference (e.g. asymptotic behavior as more data is collected, posterior contraction rates, etc.). For the first point, and under the deterministic setting, theoretical results in compressed sensing \cite{donoho2006compressed} and sparse regression can provide a path to obtain theoretical guarantees. For example, the work of Brunton {\em et al.} \cite{brunton2016discovering} discusses the effect of observation noise on the identifiability of a dynamical system \cite{schmidt2011automated, daniels2015automated}. For the latter point, one could gain theoretical insight from analyzing the properties of HMC samplers as discussed in \cite{neal2011mcmc,betancourt2017geometric}, but in practice we may only rely on appropriate tests of statistical convergence such as the Geweke analysis and the Gelman Rubin tests \cite{gelman2013bayesian, cowles1996markov}, similarly to the results presented in section \ref{sec:Results}\ref{sec:Damped}\ref{sec:diagnostic}.

\section*{Acknowledgements}
This work received support from DOE grant DE-SC0019116, AFOSR grant FA9550-20-1-0060, and DOE-ARPA grant 1256545. The human motion data used in this project was obtained from \url{mocap.cs.cmu.edu}. The human motion database was created with funding from NSF EIA-0196217.

\bibliographystyle{unsrt}
\bibliography{main.bib}

\end{document}